\documentclass{jmlr} 

\usepackage{longtable}

\usepackage{booktabs}

\jmlrvolume{vol}
\jmlryear{2012} \usepackage{bm}
\usepackage{amssymb}
\usepackage{graphicx}
\usepackage{amssymb}
\usepackage{amsmath}
\usepackage{graphicx}
 \usepackage{float}
\newcommand{\x}{\bm{x}}
\newcommand{\y}{\bm{y}}
\newcommand{\s}{\bm{s}}

\newcommand{\f}{\mathbf{f}}
\newcommand{\ff}{\mathfrak{f}}
\newcommand{\cf}{\mathfrak{F}}
\newcommand{\tcf}{\widetilde{\mathcal{F}}}
\newcommand{\bta}{\pmb\eta}
\newcommand{\bpi}{\pmb\pi}
\newcommand{\bxi}{\mathbf{H}}
\newcommand{\cK}{\mathcal{K}}

\newcommand{\af}{\bar{f}}
\newcommand{\rf}{\tilde{f}}
\newcommand{\Z}{\mathbf{Z}}
\newcommand{\bH}{\mathbf{H}}
\newcommand{\F}{\mathbf{F}}
\newcommand{\G}{\mathbf{G}}

\newcommand{\tf}{\mathbf{\tilde{f}}}
\newcommand{\D}{\breve{\y}}
\newcommand{\T}{\mathrm{T}}

\newcommand{\fK}{\mathbf{C}}
\newcommand{\fL}{\bm{L}}
\newcommand{\fA}{\bm{A}}
\newcommand{\fB}{\bm{B}}
\newcommand{\fC}{\bm{C}}
\newcommand{\fQ}{\bm{Q}}
\newcommand{\pk}{\mathbf{C}}
\newcommand{\tkp}{\widetilde{\pk}_{*j}\left[\widetilde{\fK}_{jj} + \sigma_j^2\mathbb{I}_j\right]^{-1}}
\newcommand{\tkpt}{\left(\widetilde{\fK}_{jj} + \sigma_j^2\mathbb{I}_j\right)^{-1}\widetilde{\pk}_{j*}}

\def\bbbe{{\rm I\!E}} 
\def\bbbr{{\rm I\!R}} 
\newtheorem{assumption}{Assumption}
\newcommand{\ignore}[1]{}

\mainmatter  

\title{Nonparametric Bayesian Mixed-effect Model: a Sparse Gaussian Process Approach}
\author{\Name{Yuyang Wang} \Email{ywang02@cs.tufts.edu}
  \\
  \Name{Roni Khardon} \Email{roni@cs.tufts.edu}\\
  \addr
Department of Computer Science, Tufts University, Medford, MA 02155, USA}

\begin{document}
\maketitle

\begin{abstract}
Multi-task learning models using Gaussian processes (GP)
have been developed and successfully applied in various
applications. The main difficulty with this approach is the
computational cost of inference using the union of examples from all
tasks. Therefore sparse solutions, that avoid using the entire data directly and instead use a set of informative ``representatives'' are desirable. The paper investigates this problem for the grouped
mixed-effect GP model where each individual response is given by a
fixed-effect, taken from one of a set of unknown groups, plus a random
individual effect function that captures variations among individuals.
Such models have been widely used in previous work but no sparse
solutions have been developed. The paper presents the first sparse
solution for such problems, showing how the sparse approximation can
be obtained by maximizing a variational lower bound on the marginal
likelihood, generalizing ideas from single-task Gaussian processes to
handle the mixed-effect model as well as grouping.  Experiments using
artificial and real data validate the approach showing that it can
recover the performance of inference with the full sample, that it
outperforms baseline methods, and that it outperforms state of the art
sparse solutions for other multi-task GP formulations.
\end{abstract}


\section{Introduction}

In many real world problems we are interested in learning multiple tasks
while the training set for each task is quite small.  When the different
tasks are related one can learn all tasks simultaneously and aim
to get improved predictive performance by taking advantage of the common
aspects of all tasks. This general idea is known as multi-task learning and it
has been successfully investigated in several technical settings, with
applications in many areas including medical diagnosis~\citep{biimproved},
recommendation systems~\citep{dinuzzo2008client} and HIV Therapy
Screening~\citep{bickel2008multi}.

In this paper we explore Bayesian models especially using Gaussian
Processes (GP) where sharing the prior and its parameters among the
tasks can be seen to implement multi-task learning
\citep{alvarez2011kernels,bonilla2008multi,xue2007multiBB,gelman2004bayesian,yu2005learning,schwaighofer2005lgp,pillonetto2010bayesian}. Our focus is on the \emph{functional mixed-effect model}
\citep{lu2008rkh,pillonetto2010bayesian} where each task is modeled
as a sum of a fixed effect shared by all the tasks and a random
effect that can be interpreted as representing task specific
deviations.  In particular, both effects are realizations of
zero-mean Gaussian processes.  Thus, in this model, tasks share
structure through hyper-parameters of the prior and through the
fixed effect portion.  This model has shown success in several
applications, including geophysics~\citep{lu2008rkh},
medicine~\citep{pillonetto2010bayesian} and
astrophysics~\citep{wang2010shift}. One of the main difficulties with
this model, however, is computational cost, because while the number
of samples per task $N_j$ is small, the total sample size
$\sum_{j}N_{j}$ can be large, and the typical cubic complexity of GP
inference can be prohibitively large \citep{yu2005learning}.  Some
improvement can be obtained when all the input tasks share the same
sampling points, or when different tasks share many of the input
points \citep{pillonetto2009fast,pillonetto2010bayesian}.  However,
if the number of distinct sampling points is large the complexity
remains high.  For example, this is the case in~\citep{wang2010shift}
where sample points are clipped to a fine grid to avoid the high
cardinality of the example set.

The same problem, handling large samples, has been addressed in single task formalizations of GP, where several approaches for so-called sparse solutions have been developed
\citep{rasmussen2005gaussian,Seeger03,Snelson06,titsias2008variational}.
These
methods approximate the GP with $m\ll N$ support
variables (or inducing variables, pseudo inputs) $\mathcal{X}_m$ and their corresponding function values $\f_m$ and perform inference using this set.

In this paper, we develop a sparse solution for multi-task learning
with GP in the context of the functional mixed effect model.
Specifically, we extend the approach of
\cite{titsias2008variational} and develop a variational
approximation that allows us to efficiently learn the shared
hyper-parameters and choose the sparse pseudo samples. In addition,
we show how the variational approximation can be used to perform
prediction efficiently once learning has been performed. Our
approach is particularly useful when individual tasks have a small
number of samples, different tasks do not share sampling points, and
there is a large number of tasks. Our experiments, using artificial
and real data, validate the approach showing that it can recover the
performance of inference with the full sample, that it performs
better than simple sparse approaches for multi-task GP, and that for some applications it significantly outperforms alternative sparse multi-task GP formulation~\citep{alvarez2011computationally}.

To summarize, our contribution is threefold. First we introduce the first
sparse solution for the multi-task GP in mixed-effect model. Second, we develop a variational
model-selection approach for the proposed sparse model. Finally we evaluate
the algorithm and several baseline approaches for multi-task GP, showing that
the proposed method performs well.

This paper is organized as follows. Section 2 reviews the mixed-effect GP model and its direct inference. Section 3 develops the variational inference and model selection
for the sparse mixed-effect GP model. Section 4 shows how to extend the sparse solution to the grouped mixed-effect GP model. We discuss related work in Section 5 and demonstrate the performance of the proposed approach using
three datasets in Section 6. Section~7 concludes with a summary and directions for future work.

\section{Mixed-effect GP for Multi-task Learning}
In this section and the next one, we develop the mixed-effect model and its sparse solution without considering grouping. The model and results are extended to include grouping in Section 4. Consider a set of $M$ tasks where the data for the $j$th task is given
by $\mathcal{D}^j=\{(\x_i^j,y_i^j)\}, i = 1,2,\cdots,N_j$. Multi-task learning
aims to learn all tasks simultaneously, taking the advantage of
common aspects of different tasks. In this paper, given data
$\D=\{\mathcal{D}^j\}$, we are interested in learning the nonparametric Bayesian
mixed-effect model and using the model to perform inference. The
model captures each task $f^j$ as a sum of an average effect
function and an individual variation specific to the $j$th
task. More precisely~\citep{pillonetto2010bayesian}:
\begin{assumption}
  For each $j$ and $\x\in\mathcal{X}\subset\bbbr^d$,
  \begin{equation}
  \label{eqn:bmix}
    f^j(\x) = \bar{f}(\x) + \tilde{f}^j(\x), \quad j=1,\cdots, M
  \end{equation}
  where $\bar{f}$ and $\{\tilde{f}^j\}$ are zero-mean
  Gaussian processes. In addition, $\af$ and the set of $\{\rf^j\}$ are assumed to be
  mutually independent with covariance functions $K(\cdot, \cdot)$ and $\widetilde{K}(\cdot, \cdot)$ respectively.
\end{assumption}
Assumption 1 implies that for $j, l \in \{1,\cdots, M\}$, the
following holds:
\begin{equation}
\begin{split}
\textbf{Cov}[f^j(\s), f^l(\bm{t})]
& = \textbf{Cov}[\bar{f}(\s), \bar{f}(\bm{t})] + \delta_{jl}\cdot\textbf{Cov}[\tilde{f}(\s), \tilde{f}(\bm{t})]
\end{split}
\end{equation}
where $\delta_{jl}$ is the Kronecker delta function. Let
$\breve{\x}$ be the concatenation of the examples from all tasks
$\breve{\x}=(\x_i^j)$, and similarly let $\breve{\y} = (\y_i^j)$,
where $ i = 1,2,\cdots, N_j, j = 1,2,\cdots, M$ and $N = \sum_jN_j$.
It can easily been seen that, for any $j\in\{1,\cdots, M\}$ and new
input $\x^*$ for task $j$, we have
\begin{equation}
\label{eqn:fi}
  \begin{bmatrix}
   f^j(\x^*)\\
   \breve{\y}
  \end{bmatrix}
  \sim
  \mathcal{N}\left(\mathbf{0}, \begin{bmatrix}
    \fK^\dag(\x^*, \x^*) & \fK^\dag(\x^*, \breve{\x})\\
    \fK^\dag(\breve{\x}, \x^*) & \fK^\dag(\breve{\x}, \breve{\x}) + \sigma^2\mathbb{I}
  \end{bmatrix}
  \right)
\end{equation}
where the covariance matrix $\fK^\dag$ is given by
 \[ \fK^\dag((\x_i^j),(\x_k^l)) = K(\x_i^j,\x_k^l) + \delta_{jl}\cdot\widetilde{K}((\x_i^j,\x_k^l).\]

From~(\ref{eqn:fi}) we can extract the marginal distribution
$\Pr(\breve{\y})$ where
\begin{equation}
\label{eq:fullMlikelihood}
  \breve{\y}|\breve{\x} \sim \mathcal{N}(\mathbf{0}, \fK^\dag(\breve{\x}, \breve{\x}) +
  \sigma^2\mathbb{I}),
\end{equation}
which can be used for model selection, that is, learning the
hyper-parameters of the GP. (\ref{eqn:fi}) also provides the predictive distribution where
\begin{equation}
\label{eq:fullpredictive}
\begin{split}
  \bbbe(f^j(\x^*)) &= \fK^\dag(\x^*, \breve{\x})(\fK^\dag(\breve{\x}, \breve{\x}) + \sigma^2\mathbb{I})^{-1}\breve{\y}\\
  \textbf{Cov}(f^j(\x^*)) &= \fK^\dag(\x^*, \x^*) - \fK^\dag(\x^*, \breve{\x})
  (\fK^\dag(\breve{\x}, \breve{\x}) + \sigma^2\mathbb{I})^{-1}\fK^\dag(\breve{\x}, \x^*).
\end{split}
\end{equation}
This works well in that sharing the information improves predictive
performance but, as the number of tasks grows, the dimension $N$
increases leading to slow inference scaling as $\mathcal{O}(N^3)$. In other words, even though
each task may have a very small sample, the multi-task inference problem
becomes infeasible when the number of tasks is large.


In single task GP regression, to reduce the computational cost,
several sparse GP approaches have been proposed
\citep{rasmussen2005gaussian,Seeger03,Snelson06,titsias2008variational}.
In general, these
methods approximate the GP with a small number $m \ll N$ of support
variables and
perform inference using this subset and the corresponding function values $\f_m$.
Different approaches differ in how they choose the support variables
and the simplest approach is to choose a random subset of the given
data points. Recently, \cite{titsias2008variational}
introduced a sparse method based on variational inference using a
set $\mathcal{X}_m$ of inducing samples, which are different from
the training points.
In this approach, the sample points $\mathcal{X}_m$ are chosen to maximize
a variational lower bound on the marginal likelihood, therefore providing a
clear methodology for the choice of the support set. Following their idea,~\cite{alvarez2010efficient} proposed the variational inference for sparse convolved multiple output GPs.

In this paper we extend this approach to provide a sparse solution for the aforementioned model as well as generalizing it to the Grouped mixed-effect GP model~\citep{wang2010shift}. As in the case of sparse methods for single task GP, the key idea is to introduce a small set of $m$ auxiliary inducing sample points
$\mathcal{X}_m$ and base the learning and inference on these points.
For the multi-task case, each $\tilde{f}^j(\cdot)$ is specific to
the $j$th task. Therefore, it makes sense to induce values only for
the fixed-effect portion $\f_m=\bar{f}(\mathcal{X}_m)$. The details
of this construction are developed in the following sections.

\section{Sparse mixed-effect GP Model}
\label{sec:sinfer}
In this section, we develop a sparse solution for the mixed-effect model without group effect. The model is simpler to analyze and apply, and it thus provides a good introduction to the results developed in the next section for the grouped model.

\subsection{Variational Model Selection}
In this section we specify the sparse model,
and show how we can learn the hyper-parameters and the
inducing variables using the sparse model.
As mentioned above, we introduce
auxiliary inducing sample points $\mathcal{X}_m$
and hidden variables
$\f_m=\bar{f}(\mathcal{X}_m)$.
Let $\f^{j}=\bar{f}(\x^j)\in\bbbr^{N_{j}}$ and $\tf^{j}=\tilde{f}(\x^j)\in\bbbr^{N_{j}}$
denote the values of the two functions at
$\x^{j}$.
In addition let $\fK_{*j}=\fK(\x^*, \x^j)$, $\fK_{jj}=
\fK(\x^j, \x^j)$ and $\fK_{mm}=\fK(\mathcal{X}_m,
\mathcal{X}_m)$, and similarly for
$\widetilde{\fK}_{*j}, \widetilde{\fK}_{jj}, \widetilde{\fK}_{mm}$.


To learn the hyper-parameters we wish to maximize the marginal
likelihood $\Pr(\breve{\y})$ where $\breve{\y}$ is all the
observations. In the following we develop a variational lower bound
for this quantity. To this end, we need the complete data likelihood
and the variational distribution.
\begin{itemize}
  \item The complete data likelihood $\Pr(\{\y^j\}, \{\f^{j},
\tf^{j}\}, \f_m)$ is given by:
\begin{displaymath}
\begin{split}
 &\Pr(\{\y^j\}|\{\f^{j}, \tf^{j}\})\Pr(\{\tf^j\})\Pr(\{\f^j\}|\f_m)\Pr(\f_m) \\
  &\quad= \left[ \prod_{j=1}^M\Pr(\y^j|\f^j, \tf^j)\Pr(\tf^j)\right] \Pr(\{\f^j\}|\f_m)\Pr(\f_m).
\end{split}
\end{displaymath}
  \item We approximate the posterior $\Pr(\{\f^{j}, \tf^{j}\},
\f_m | \{\y^j\}) $ on the hidden variables by
\begin{equation}
\label{eqn:varform1}
q(\{\f^{j}, \tf^{j}\}, \f_m) = \left[ \prod_{j=1}^M\Pr(\tf^j|\f^j, \y^j)\right]
      \Pr(\{\f^j\}|\f_m)\phi(\f_m)
\end{equation}
which extends the variational form used by
\cite{titsias2008variational} to handle the individual variations as well as the multiple tasks.
One can see that the variational distribution is not completely in
free form. Instead, $q(\cdot)$ preserves the exact form of
$\Pr(\tf^j|\f^j, \y^j)$ and in using $\Pr(\{\f^j\}|\f_m)$ it implicitly
assumes that $\f_m$ is a sufficient statistic for $\{\f^j\}$. The free
form $\phi(\f_m)$ corresponds to $\Pr(\f_m|\D)$ but allows it to
diverge from this value to compensate for the assumption that $\f_m$
is sufficient.
Notice that we
are not making any assumption about the sufficiency of $\f_m$ in the
generative model and the approximation is entirely due to
the variational distribution. An additional assumption is added later to derive a simplified form of the predictive distribution.
\end{itemize}

With the two ingredients ready, the variational lower bound
~\citep{Jordan99,bishop2006pattern}, denoted as
$F_V(\mathcal{X}_m,\phi)$, is given by:
\begin{displaymath}
    \begin{split}
      \Pr(\breve{\y}) &\geqslant F_V(\mathcal{X}_m,\phi) \\
      &= \int q(\{\f^{j}, \tf^{j}\}, \f_m)\times \log\left[\frac{\Pr(\{\y^j\}, \{\f^{j}, \tf^{j}\}, \f_m)}{q(\{\f^{j}, \tf^{j}\}, \f_m)}\right]d\{\f^j\}d\{\tf^j\}d\f_m \\
      &\ = \int\left[ \prod_{j=1}^M\Pr(\tf^j|\f^j, \y^j)\right]
      \Pr(\{\f^j\}|\f_m)\phi(\f_m)\\
      &\ \ \ \times \log\left[\prod_{l=1}^M\frac{\Pr(\y^l|\f^l,\tf^l)\Pr(\tf^l)}{\Pr(\tf^l|\f^l, \y^l)}
      \cdot \frac{\Pr(\f_m)}{\phi(\f_m)}\right]d\{\f^j\} d\{\tf^j\} d\f_m\\
      &\ = \int\phi(\f_m)\left\{\log G(\f_m, \mathcal{Y}) +
      \log\left[\frac{\Pr(\f_m)}{\phi(\f_m)}\right]\right\}d\f_m.
    \end{split}
\end{displaymath}
The inner integral denoted as $\log G(\f_m, \mathcal{Y})$ is
\begin{equation}
  \label{eqn:logG1}
  \begin{split}
    &\int\left[ \prod_{j=1}^M\Pr(\tf^j|\f^j, \y^j)\right] \Pr(\{\f^j\}|\f_m)\times\sum_{l=1}^M\log\left[\frac{\Pr(\y^l|\f^l,\tf^l)\Pr(\tf^l)}{\Pr(\tf^l|\f^l, \y^l)}\right]d\{\f^j\} d\{\tf^j\}\\
    &\ = \sum_{j=1}^M\int \Pr(\tf^j|\f^j, \y^j)\Pr(\f^j|\f_m)\times \log\left[\frac{\Pr(\y^j|\f^j,\tf^j)\Pr(\tf^j)}{\Pr(\tf^j|\f^j, \y^j)}\right]d\f^j d\tf^j
  \end{split}
\end{equation}
where the second line holds because in the sum indexed by $l$ all
the product measures
\[\prod_{j=1, j\neq l}^M\Pr(\tf^j|\f^j,
\y^j)\Pr(\{\f^n\}_{n\neq l}|\f_m, \f_l)d\{\f^j\} d\{\tf^j\},\]  are
integrated to 1, leaving only the $j$-th integral. 
In subsection 3.1 we show that
\begin{equation}
\label{eqn:Gfm1}
\begin{split}
  \log G(\f_m, \mathcal{Y}) &= \sum_{j=1}^m\Bigg[\log\left[\mathcal{N}(\y^{j}|\boldsymbol{\alpha}_j,
  \widehat{\fK}_{jj})\right]- \frac{1}{2}\textbf{Tr}\left[(\fK_{jj} - \fQ_{jj})[\widehat{\fK}_{jj}]^{-1}\right]\Bigg]
\end{split}
\end{equation}
where $\boldsymbol{\alpha}_j=\fK_{jm}\fK_{mm}^{-1}\f_m$,
$\widehat{\fK}_{jj}=\sigma_j^2\mathbb{I} + \widetilde{\fK}_{jj}$, and
$\fQ_{jj} = \fK_{jm}\fK_{mm}^{-1}\fK_{mj}$. Thus we have
\begin{equation}
\label{eqn:bound}
\begin{split}
   F_V(\mathcal{X}_m, \phi)&= \int\phi(\f_m)\left[\log G(\f_m, \mathcal{Y}) + \log\left[\frac{\Pr(\f_m)}{\phi(\f_m)}\right]\right]d\f_m\\
  &=\int\phi(\f_m)\log\left[\frac{\prod_j\left[\mathcal{N}(\y^j|\boldsymbol{\alpha}_j,
  \widehat{\fK}_{jj})\right]\Pr(\f_m)}{\phi(\f_m)}\right]d\f_m\\
&\qquad\qquad\qquad-\frac{1}{2}\sum_{j=1}^M\textbf{Tr}\left[(\fK_{jj} -
\fQ_{jj})[\widehat{\fK}_{jj}]^{-1}\right].
\end{split}
\end{equation}


Let $\bm{v}$ be a random variable and $g$ any function, then by
Jensen's inequality $\bbbe[\log g(\bm{v})]\leq
\log\bbbe[g(\bm{v})]$. Therefore, the best lower bound we can derive
from ~(\ref{eqn:bound}), if it is achievable, is the case where
equality holds in Jensen's inequality. In subsection 3.2 we
show that $\phi(\f_m)$ can be chosen to obtain equality, and therefore, the variational lower
bound is
\begin{displaymath}
\begin{split}
  F_V(\mathcal{X}_m, \phi) &= \log \int \prod_j\left[\mathcal{N}(\y^j|\boldsymbol{\alpha}_j,
  \widehat{\fK}_{jj})\right]\Pr(\f_m)d\f_m- \frac{1}{2}\sum_{j=1}^M\text{Tr}\left[(\fK_{jj} - Q_{jj})[\widehat{\fK}_{jj}]^{-1}\right].
\end{split}
\end{displaymath}
Evaluating the integral by marginalizing out $\f_m$ and recalling that $\D$ is the concatenation of the $\y^j$, we get
\begin{equation}
\label{eqn:lb}
\begin{split}
  F_V(\mathcal{X}_m, \boldsymbol{\theta},
  \boldsymbol{\tilde{\theta}})&= 
   \log\left[\mathcal{N}(\D|\mathbf{0}, \mathbf\Lambda_m\fK_{mm}^{-1}\mathbf\Lambda_m^T + \widehat{\fK}^m)\right]- \sum_{j=1}^m\bigg[\frac{1}{2}\textbf{Tr}\left[(\fK_{jj} - \fQ_{jj})[\widehat{\fK}_{jj}]^{-1}\right]\bigg]
  \end{split}
\end{equation}
where
\[
\mathbf\Lambda_m = \begin{pmatrix}
  \fK_{1m}\\
  \fK_{2m}\\
  \vdots\\
  \fK_{Mm}
\end{pmatrix} \qquad \text{and}\qquad
\widehat{\fK}^m = \bigoplus_{j=1}^M \widehat{\fK}_{jj} = \begin{pmatrix}
  \widehat{\fK}_{11} & & &\\
  & \widehat{\fK}_{22} & & \\
  & & \ddots & \\
  & & & \widehat{\fK}_{MM}
\end{pmatrix}.
\]
Thus, we have explicitly written the parameters that can be chosen
to further optimize the lower bound, namely the support inputs
$\mathcal{X}_m$, and the hyper-parameters $\boldsymbol\theta$ and
$\boldsymbol{\tilde{\theta}}$ in ${K}$ and $\widetilde{{K}}$
respectively. By calculating derivatives of ~(\ref{eqn:lb}) we
can optimize the lower bound using a gradient based method. In the
experiments in this paper, we use stochastic gradient descent (SGD),
which works better than the conjugate gradient (CG) in this scenario
where the number of tasks is large.
%
%
~\cite{titsias2008variational} outlines
methods that can be used when gradients are not useful.

\subsubsection{Evaluating $\log G(\f_m, \mathcal{Y})$}

Consider the $j$-th element in the sum of ~(\ref{eqn:logG1}):
\begin{displaymath}
  \begin{split}
    &\widehat{G}_j(\f^j, \y^j) =\int\Pr(\tf^j|\f^j, \y^j)\Pr(\f^j|\f_m)\log\left[\frac{\Pr(\y^j|\f^j,\tf^j)\Pr(\tf^j)}{\Pr(\tf^j|\f^j,\y^j)}\right]d\f^j d\tf^j\\
    &\ =\int\Pr(\tf^j|\f^j, \y^j)\Pr(\f^j|\f_m)\\
    &\qquad \ \times\log\left[\frac{\Pr(\tf^j|\y^j,\f^j)\Pr(\y^j|\f^j)}{\Pr(\tf^j|\f^j)}\cdot\frac{\Pr(\tf^j)}{\Pr(\tf^j|\f^j,\y^j)}\right]d\f^j d\tf^j\\
    &=\int\Pr(\f^j|\f_m)\log\left[\Pr(\y^j|\f^j)\right]\left(\int\Pr(\tf^j|\f^j, \y^j)d\tf^j\right) d\f^j\\
    &\ = \int\Pr(\f^j|\f_m)\log\left[\Pr(\y^j|\f^j)\right]d\f^j
    = \bbbe_{[\f^j|\f_m]}\log\left[\Pr(\y^j|\f^j)\right]\\
  \end{split}
\end{displaymath}
where the third line holds because of the independence between $\tf^j$
and $\f^j$. We next show how this expectation can be evaluated.
This is more complex than the single-task case because of the coupling of the fixed-effect and the random effect.

Recall that \[
  \Pr(\f^j|\f_m) = \mathcal{N}(\f^j|\fK_{jm}\fK_{mm}^{-1}\f_m, \fK_{jj}-\fK_{jm}{\fK}_{mm}^{-1}{\fK}_{mj})
\] and \[\y^j|\f^j \sim\mathcal{N}(\f^j,\widehat{\fK}_{jj})\]
where $\widehat{\fK}_{jj}=\sigma^2\mathbb{I} +
\widetilde{\fK}_{jj}$. Denote $\widehat{\fK}_{jj}^{-1}=\fL^\T\fL$
where $\fL$ can be chosen as its Cholesky decomposition, we have
\begin{displaymath}
\begin{split}
   \log\left[\Pr(\y^j|\f^j)\right] &= - \frac{1}{2}(\y^j-\f^j)^T\widehat{\fK}_{jj}^{-1}(\y^j-\f^j) + \log\left[(2\pi)^{-\frac{N_j}{2}}\right] + \log\left[|\widehat{\fK}_{jj}|^{-\frac{1}{2}}\right]\\
   &= -\frac{1}{2}(\fL\y^j - \fL\f^j)^T(\fL\y^j - \fL\f^j) + \log\left[(2\pi)^{-\frac{N_j}{2}}\right] + \log\left[|\widehat{\fK}_{jj}|^{-\frac{1}{2}}\right].
\end{split}
\end{displaymath}
Notice that \[\Pr(\fL\f^j|\f_m) =
\mathcal{N}(\fL\fK_{jm}\fK_{mm}^{-1}\f_m, \fL(\fK_{jj} -
\fQ_{jj})\fL^\T)\] where $\fQ_{jj} = \fK_{jm}\fK_{mm}^{-1}\fK_{mj}$.
Recall the fact that for $\x\sim\mathcal{N}(\boldsymbol{\mu}, \mathbf{\Sigma})$
and a constant vector $\bm{a}$, we have $\bbbe[\|\bm{a} - \x\|^2] =
\|\bm{a} - \boldsymbol{\mu}\|^2 + \textbf{Tr}(\mathbf{\Sigma})$.
Thus,
\begin{displaymath}
\begin{split}
  &\bbbe_{[\f^j|\f_m]}\log\left[\Pr(\y^j|\f^j)\right]  =-\frac{1}{2}\|\fL\y^j - \fL\fK_{jm}\fK_{mm}^{-1}\f_m\|^2
  \\
  &\ \qquad -\frac{1}{2}\textbf{Tr}(\fL(\fK_{jj} - \fQ_{jj})\fL^T)+ \log\left[(2\pi)^{-\frac{N_j}{2}}\right] + \log\left[|\widehat{\fK}_{jj}|^{-\frac{1}{2}}\right]\\
  &= \Bigg\{-\frac{1}{2}\left[\y - \fK_{jm}\fK_{mm}^{-1}\f_m\right]^T\left(L^TL\right)\left[\y - \fK_{jm}\fK_{mm}^{-1}\f_m\right]\\
  &\qquad  + \log\left[(2\pi)^{-\frac{N_j}{2}}\right]+ \log\left[|\fK_{jj}|^{-\frac{1}{2}}\right]\Bigg\} - \frac{1}{2}\text{Tr}\left[L(\fK_{jj} - \fQ_{jj})L^T\right]\\
  &\ = \log\left[\mathcal{N}(\y^j|\boldsymbol{\alpha}_j, \widehat{\fK}_{jj})\right] - \frac{1}{2}\textbf{Tr}\left[(\fK_{jj} - \fQ_{jj})\widehat{\fK}_{jj}^{-1}\right]
\end{split}
\end{displaymath}
where $\boldsymbol{\alpha}_j=\fK_{jm}\fK_{mm}^{-1}\f_m$.
Finally, calculating
$\sum_j \widehat{G}_j(\f^j, \y^j)$
we get~(\ref{eqn:Gfm1}).

\subsubsection{Variational distribution $\phi^{*}(\f_{m})$}

For equality to hold in Jensen's inequality, the
function inside the $\log$ must be constant.
In our case this is easily achieved because $\phi(\f_m)$ is a free parameter, and we can set
\begin{displaymath}
\left[\frac{\prod_j\left[\mathcal{N}(\y^j|\boldsymbol{\alpha}_j,\widehat{\fK}_{jj})
\right]\Pr(\f_m)}{\phi(\f_m)}\right]\equiv c,
\end{displaymath}
yielding the bound given in~(\ref{eqn:lb}). Setting $\phi(\f_m) \propto \prod_j\left[\mathcal{N}(\y^j|\boldsymbol{\alpha}_j,\widehat{\fK}_{jj})\right]\Pr(\f_m)$
yields the form of the optimal
variational distribution
\begin{displaymath}
\begin{split}
\phi^*(\f_m)
&\propto \prod_j\left[\mathcal{N}(\y^j|\boldsymbol{\alpha}_j,\widehat{\fK}_{jj})\right]\Pr(\f_m)\\
&\propto \exp\Bigg\{-\frac{1}{2}\f_m^T\left[\fK_{mm}^{-1}\boldsymbol{\Phi}\fK_{mm}^{-1}\right]\f_m +
\f_m^\T\left(\fK_{mm}^{-1}\sum_j\fK_{mj}\left[\widehat{\fK}_{jj}\right]^{-1}\y^j\right)\Bigg\},
\end{split}
\end{displaymath}
from which we observe that $\phi^*(\f_m)$ is
\begin{equation}
\label{eqn:vardis1}
\begin{split}
\mathcal{N}\Bigg(\f_m\Bigg|\fK_{mm}\boldsymbol{\Phi}^{-1}\sum_j\fK_{mj}[\widehat{\fK}_{jj}]^{-1}\y_j,
\fK_{mm}\boldsymbol{\Phi}^{-1}\fK_{mm}\Bigg)
\end{split}
\end{equation}
where
$
\boldsymbol{\Phi} = \fK_{mm} +
\sum_j\fK_{mj}[\widehat{\fK}_{jj}]^{-1}\fK_{jm}.
$
 Notice that by choosing the number of tasks to be 1 and the random effect to be a noise process,
 i.e. $\widetilde{K}(s, t) = \sigma^2\delta(s, t)$,  (\ref{eqn:lb}) and  (\ref{eqn:vardis})
 are exactly the variational lower bound and the corresponding variational distribution in~\citep{titsias2008variational}.

\subsection{Prediction using the Variational Solution}
\label{sec:sing_predict}
Given any task $j$, our goal is to calculate the predictive
distribution of $f^j(\x^*) = \bar{f}(\x^*) + \tilde{f}^j(\x^*)$ at
some new input point $\x^*$. As described before, the full inference
is expensive and therefore we wish to use the variational
approximation for the prediction as well. The key assumption is that
$\f_m$ contains as much information as $\D$ in terms of making
prediction for $\af$. To start with, it is easy to see that the
predictive distribution is Gaussian and that it satisfies
\begin{equation}
\label{eq:predictj1}
  \begin{split}
    \bbbe[f^j(\x^*)|\D] &= \bbbe[\bar{f}(\x^*)|\D] +
    \bbbe[\tilde{f}^j(\x^*)|\D]\\
    \textbf{Var}[f^j(\x^*)|\D] &= \textbf{Var}[\af(\x^*)|\D] + \textbf{Var}[\rf^j(\x^*)|\D] + 2\textbf{Cov}[\af(\x^*)\rf^j(\x^*)|\D].
  \end{split}
\end{equation}
The above equation is more complex than the predictive distribution for
single-task sparse GP because of the coupling induced by
$\af(\x^*)\rf^j(\x^*)|\D$. We next show how this can be calculated via
conditioning.

To calculate the terms in~(\ref{eq:predictj1}), three parts are needed, i.e., $\Pr(\af(\x^*)|\D)$, $\Pr(\rf(\x^*)|\D)$ and $\textbf{Cov}[\af(\x^*)\rf_j(\x^*)|\D]$.
Using the assumption of the variational form given in~(\ref{eqn:varform1}), we have the following facts,

\begin{enumerate}
\item $\f_m | \D \sim \phi^*(\f_m) = \mathcal{N}(\boldsymbol\mu, \fA)$ where $\boldsymbol\mu$ and $\fA$ are given in~(\ref{eqn:vardis1}).

\item $\f_m$ is sufficient for $\{\f^j\}$, i.e.
$\Pr(\{\f^{j}\}|\f_m, \D)= \Pr(\{\f^{j}\}|\f_m)$. Since we are
interested in prediction for each task separately, by marginalizing
out $\f^l, l\neq j$, we also have
$\Pr(\f^{j}|\f_m, \D)= \Pr(\f^{j}|\f_m)$ and
\begin{equation}\label{eqn:fjm}\f^{j}|\f_m, \D \sim \mathcal{N}\left(\fK_{jm}\fK_{mm}^{-1}\f_m, \fK_{jj}-\fK_{jm}\fK_{mm}^{-1}\fK_{mj}\right).\end{equation}

\item For $\rf^j(\x^*)$ we can view $\y^j -\f^j$ as noisy realizations
from the same GP as $\rf^j(\x^j)$ and therefore
\begin{equation}
\label{eqn:ftjx1}
\begin{split}
    \rf^j(\x^*)|\f^j, \D &\sim \mathcal{N}\Bigg(\tkp\left[\y^j -
    \f^j\right], \widetilde{\fK}_{**}- \tkp\widetilde{\fK}_{j*}\Bigg).
\end{split}
\end{equation}

\end{enumerate}

In order to obtain a sparse form of the predictive distribution we need to make an additional assumption.
\begin{assumption}
  We assume that $\f_m$ is sufficient for $\af(\x^*)$,
i.e., \[\Pr(\af(\x^*)|\f_m, \D) = \Pr(\af(\x^*)|\f_m),\] implying that
\begin{equation}
\label{eqn:fbars1}
\af(\x^*)|\f_m, \D \sim \mathcal{N}\left(\fK_{*m}\fK_{mm}^{-1}\f_m,
  \fK_{**}-\fK_{*m}\fK_{mm}^{-1}\fK_{m*}\right).\end{equation}
  \end{assumption}
The above set of conditional distributions also imply that
$\af(\x^*)$ and $\rf^j(\x^*)$ are independent given $\f_m$ and $\D$.

To evaluate (\ref{eq:predictj1}), we have the following
\begin{itemize}
  \item We can easily get $\Pr(\af(\x^*)|\D)$ by marginalizing out $\f_m|\D$ in~(\ref{eqn:fbars1}),
  \begin{displaymath}
    \Pr(\af(\x^*)|\D) = \int\Pr(\af(\x^*)|\f_m)\phi^*(\f_m)d\f_m
  \end{displaymath}
yielding
  \begin{equation}
  \label{eqn:Rpfix}
  \begin{split}
    \af(\x^*)|\D &\sim \mathcal{N}\Bigg(\fK_{*m}\fK_{mm}^{-1}\mu,\fK_{**}-\fK_{*m}\fK_{mm}^{-1}\fK_{m*} + \fK_{*m}\fK_{mm}^{-1}A\fK_{mm}^{-1}\fK_{m*}\Bigg).
  \end{split}
  \end{equation}
  \item Similarly, we can obtain $\Pr(\rf(\x^*)|\D)$ by first calculating $\Pr(\f^j|\D)$ by marginalizing out $\f_m|\D$ in~(\ref{eqn:fjm}) and then marginalizing out $\f^j|\D$ in~(\ref{eqn:ftjx1}), as follows.
First we have $\f^j|\D\sim \mathcal{N}(\fK_{jm}\fK_{mm}^{-1}\mu, \fB)$ where
\begin{displaymath}
\begin{split}
\fB &=\fK_{jj}-\fK_{jm}\fK_{mm}^{-1}\fK_{mj} + \fK_{jm}\fK_{mm}^{-1}A\fK_{mm}^{-1}\fK_{mj}.
\end{split}
\end{displaymath}
Next for $\Pr(\rf(\x^*)|\D)$, we have
  \begin{displaymath}
  \begin{split}
    \Pr(\rf^j(\x^*)|\D)
    &= \int\Pr(\rf^j(\x^*)|\f^j, \y^j)\Pr(\f^j|\D)d\f^j
  \end{split}
  \end{displaymath}
and marginalizing out $\f^j$, $\rf(\x^*)|\D$ can be obtained as
  \begin{equation}
  \begin{split}
     &\mathcal{N}\bigg(\tkp\left(\y^j - \fK_{jm}\fK_{mm}^{-1}\mu\right), \widetilde{\fK}_{**} - \tkp\widetilde{\fK}_{j*} \\
    &\qquad\quad+\tkp\fK_{jm}\fK_{mm}^{-1}\times B\fK_{mm}^{-1}\fK_{mj}\tkpt\bigg).
  \end{split}
  \end{equation}


\item Finally, to calculate $\textbf{Cov}[\af(\x^*)\rf^j(\x^*)|\D]$ we have
\begin{displaymath}
\begin{split}
  &\textbf{Cov}[\af(\x^*)\rf_j(\x^*)|\D]  = \bbbe\big[\bar{f}^j(\x^*)\cdot\tilde{f}^j(\x^*)|\D\big]- \bbbe\big[\af(\x^*)|\D\big]\bbbe\big[\rf(\x^*)|\D\big]
\end{split}
\end{displaymath}
where
\begin{equation}
\label{eqn:expect1}
  \begin{split}
    &\bbbe\big[\bar{f}^j(\x^*)\cdot\tilde{f}^j(\x^*)|\D\big]=
      \bbbe_{\f_m|\D}\bbbe\big[\bar{f}^j(\x^*)\cdot\tilde{f}^j(\x^*)|\f_m, \D\big]\\
      &\quad =
\bbbe_{\f_m|\D}
\left[\bbbe\big[\bar{f}^j(\x^*)|\f_m\big]\cdot\bbbe
  \big[\tilde{f}^j(\x^*)|\f_m, \y^j\big]\right]
  \end{split}
\end{equation}
where the second line holds because, as observed above, the terms
are conditionally independent. The first term
$\bbbe\left[\bar{f}^j(\x^*)|\f_m\right]$ can be obtained directly
from~(\ref{eqn:fbars1}). By marginalizing out $\f^j|\f_m$ in~(\ref{eqn:ftjx1}) such that
\[\Pr(\tilde{f}^j(\x^*)|\f_m, \y^j) = \int\Pr(\rf^j(\x^*)|\f^j, \D)\Pr(\f^j|\f_m)d\f^j,\]
we can get the second term. This yields

  \begin{equation}
    \begin{split}
      &\mathcal{N}\bigg(\tkp\left(\y^j -
      \fK_{jm}\fK_{mm}^{-1}\f_m\right),\widetilde{\fK}_{**} - \tkp\widetilde{\fK}_{j*} \\
      &\qquad+\tkp \fC\tkpt\bigg)
    \end{split}
  \end{equation}
  where $\fC = \fK_{jj}-\fK_{jm}\fK_{mm}^{-1}\fK_{mj}$.
To simplify
the notation, let $\bH = \fK_{*m}\fK_{mm}^{-1}$, $\F=
\widetilde{\fK}_{*j}\left(\widetilde{\fK}_{jj} +
\sigma_j^2\mathbb{I}_j\right)^{-1}$ and $\G=\fK_{jm}\fK_{mm}^{-1}$.
Then (\ref{eqn:expect1}) can be evaluated as
\begin{displaymath}
  \begin{split}
      &\bH\y^j\F\cdot\bbbe[\f_m] - \F\G\left(\bbbe\left[\f_m\f_m^\T|\D\right]\right)\bH^\T = \bH\y^j\F\cdot\mu - \F\G \left[\fA + \boldsymbol{\mu\mu}^\T\right] \bH^\T.
  \end{split}
\end{displaymath}

\end{itemize}
We have therefore shown how to calculate the predictive distribution in~(\ref{eq:predictj1}). The complexity of these computations is $\mathcal{O}(N_j^3 +
m^3)$ which is a significant improvement over $\mathcal{O}(N^3)$ where $N=M\times N_j$.

\section{Sparse Grouped mixed-effect GP Model (GMT-GP)}
In this section, we extend the mixed-effect GP model such that the fixed-effect functions admit a group structure. We call this \emph{Grouped mixed-effect GP model} (GMT-GP). More precisely, each task is sampled from a mixture of shared fixed-effect GPs and then adds its individual variation. We show how to perform the inference and model selection efficiently.
\subsection{Generative Model}
\begin{figure}[t]
{
        \centering
        \includegraphics[width=0.5\textwidth]{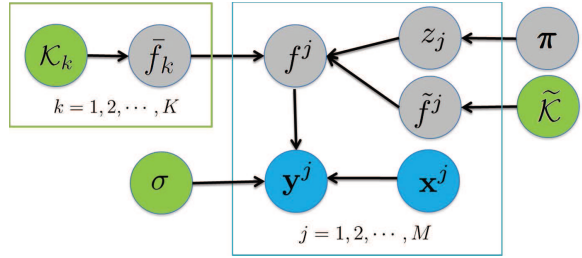}
\caption{Plate graph of the GMT-GP. Blue nodes denote observations, green ones are (hyper)parameters and the gray ones are latent variables.}
\label{fig:plate}}
\end{figure}

First, we specify the sparse GMT-GP model, and show how we
can learn the hyper-parameters and the inducing variables using this
sparse model.
\begin{assumption}
  For each $j$ and $\x\in\mathcal{X}$,
  \[
    f^j(\x) = \bar{f}_{z_j}(\x) + \tilde{f}^j(\x), \quad j=1,\cdots, M
  \]
  where $\{\bar{f}_k\}, k=1,\cdots,K$ and $\tilde{f}^j$ are zero-mean
  Gaussian processes with covariance function $\cK_k$ and $\widetilde{\cK}$, and $z_j\in\{1,\cdots,K\}$.
  In addition, $\{\bar{f}_k\}$ and $\{\tilde{f}^j\}$ are assumed to be mutually independent.
\end{assumption}
The generative process (shown in Fig.~\ref{fig:plate}) is as follows, where \textbf{Dir} and \textbf{Multi} denote the Dirichlet and the Multinominal distribution respectively.
\begin{enumerate}
  \item Draw the processes of the mean effect: $\bar{f}_k(\cdot) |\pmb\theta_k \sim \mathcal{GP}(0, \cK_k(\cdot, \cdot)),\quad k=1,2,\cdots,K$;
  \item Draw $\bpi | \pmb\alpha_0 \sim \textbf{Dir}(\pmb\alpha_0)$;
  \item For the $j$-th task (time series);
  \begin{itemize}
    \item Draw $z_j|\bpi \sim    \textbf{Multi}(\bpi)$;
    \item Draw the random effect: $\tilde{f}^j(\cdot) |\tilde{\pmb\theta} \sim \mathcal{GP}(0, \widetilde{\cK}(\cdot, \cdot))$;
    \item Draw $\y^j | z_j, f^j, \x^j, \sigma_j^2 \sim \mathcal{N}\left(f^j(\x^j),
    \sigma_j^2\cdot\mathbb{I}_j\right)$, where $f^j = \bar{f}_{z_{j}}+\tilde{f}^j$
and where to simplify the notation $\mathbb{I}_j$ stands for $\mathbb{I}_{N_j}$.
  \end{itemize}
\end{enumerate}

\subsection{Variational Model Selection}
In this section we show how to perform the learning via variational approximation. The derivation follows the same outline as in the previous section but due to the hidden variables $z_j$ that specify group membership, we have to use the variational EM algorithm. As mentioned above, for the $k$-th mixed-effect (or center), we introduce $m_k$ auxiliary inducing support variables $\mathcal{X}^k_m$ and the hidden variable $\pmb\eta_k=\bar{f}_k(\mathcal{X}^k_m)$, which is the value of $k$-th fixed-effect function evaluated at $\mathcal{X}^k_m$.

Let $\ff_k = \bar{f}_k(\breve{\x})\in\bbbr^{N}$ denote the function values of the $k$-th mean effect so that
$\f^j_k = \bar{f}_k(\x^j)\in\bbbr^{N_{j}}$ is the sub-vector of $\ff_{k}$ corresponding to the $j$-th task.
Let $\tf^{j}=\tilde{f}(\x^j)\in\bbbr^{N_{j}}$ be the values of the random effect at $\x^{j}$. Denote the collection of the hidden
variables as $\cf = \{\ff_k\}, \tcf = \{\tf^j\}, \bxi=\{\bta_k\}$,
$\Z=\{z_j\}$, and $\bpi$. In addition let $\pk^k_{*j}=\cK_k(\x^*, \x^j)$, $\fK^k_{jj}= \cK_k(\x^j, \x^j), \fK_{jk} = \cK_k(\x^j, \mathcal{X}^k_m)$ and $\fK_{kk}=\cK_k(\mathcal{X}^k_m, \mathcal{X}^k_m)$, and
similarly $\widetilde{\pk}_{*j} = \widetilde{\cK}(\x^*, \x^j), \widetilde{\fK}_{jj} = \widetilde{\cK}(\x^j, \x^j)$ and $\widehat{\fK}_{jj} = \widetilde{\fK}_{jj} + \sigma_j^2\mathbb{I}_j$ where $\mathbb{I}_j$ stands for $\mathbb{I}_{N_j}$.

To learn the hyper-parameters we wish to maximize the marginal
likelihood $\Pr(\D)$ where $\D$ is all the
measurements. In the following we develop a variational lower bound
for this quantity. To this end, we need the complete data likelihood
and the variational distribution.
The complete data likelihood is given by
  \begin{equation}
  \label{eqn:complike}
    \Pr(\D, \cf, \tcf, \bxi, \Z, \bpi) = \Pr(\D| \cf, \tcf, \Z)\Pr(\cf|\bxi)\Pr(\Z|\bpi)\Pr(\bpi)\Pr(\tcf)\Pr(\bxi)
   \end{equation}
  where
  \begin{displaymath}
    \begin{split}
      &\Pr(\bxi) = \prod_{k = 1}^K\Pr(\bta_k),\qquad \Pr(\tcf) = \prod_{j = 1}^M\Pr(\tf^j),\\
      &\Pr(\bpi) = \textbf{Dir}(\bpi | \alpha_0),\qquad \Pr(\Z|\bpi) = \prod_{j=1}^M\prod_{k = 1}^K\pi_k^{z_{jk}}\\
      &\Pr(\cf|\bxi) = \prod_{k=1}^K\Pr(\ff_k|\bta_k), \quad \Pr(\D| \cf, \tcf, \Z) = \prod_{j=1}^M\prod_{k=1}^K\left[\Pr(\y^j|\tf^j, \ff_k)\right]^{z_{jk}} \\
    \end{split}
  \end{displaymath}
where, as usual $\{z_{jk}\}$ represent $z_j$ as a unit vector.

We approximate the posterior $\Pr(\cf, \tcf, \bxi, \Z, \bpi | \D) $ on the hidden variables using
  \begin{equation}
  \label{eqn:varform}
    q(\cf, \tcf, \bxi, \Z, \bpi) = q(\cf, \tcf, \bxi| \Z)q(\Z)q(\bpi)
  \end{equation}
  where
  \[
  \begin{split}
    q(\cf, \tcf, \bxi| \Z) &= \Pr(\tcf|\cf, \Z,\D)\Pr(\cf|\bxi)\mathbf{\Phi}(\bxi)\\
    &=\prod_{j=1}^M\prod_{k=1}^K\left[\Pr(\tf^j|\ff_k, \y^j)\right]^{z_{jk}} \prod_{k=1}^K\Pr(\ff_k|\bta_k)\phi(\bta_k).
  \end{split}
  \]
This extends the variation form of the previous section. Our use of $\ff_k$ as the complete set of observation when the true group is $k$ makes for convenient notation of simplifying the derivation.

The variational lower bound, denoted as
$F_V$, is given by:
\begin{displaymath}
    \begin{split}
      \Pr(\D) \geqslant F_V &= \int q(\cf, \tcf, \bxi, \Z, \bpi)\times \log\left[\frac{\Pr(\D, \cf, \tcf, \bxi, \Z, \bpi)}{q(\cf, \tcf, \bxi, \Z, \bpi)}\right]d\cf\, d\tcf\, d\bxi\, d\Z\, d\bpi\\
      &= \int q(\bpi)q(\Z)q(\cf, \tcf, \bxi|\Z)\\
      &\quad \times \log\left[\frac{\Pr(\D| \cf, \tcf, \Z)\Pr(\cf|\bxi)\Pr(\Z|\bpi)\Pr(\pi)\Pr(\tcf)\Pr(\bxi)}{q(\cf, \tcf, \bxi| \Z)q(\Z)q(\bpi)}\right]d\cf d\tcf d\bxi d\Z d\bpi\\
      &= \int q(\Z)q(\bpi)\log\left[\frac{\Pr(\bpi)\Pr(\Z|\bpi)}{q(\Z)q(\bpi)}\right]d\bpi d\Z \\
      &\quad +
      \int q(\Z)q(\cf, \tcf, \bxi|\Z)\log\left[\frac{\Pr(\D| \cf, \tcf, \Z)\Pr(\cf|\bxi)\Pr(\tcf)\Pr(\bxi)}{q(\cf, \tcf, \bxi| \Z)}\right]d\cf d\tcf d\bxi d\Z
    \end{split}
\end{displaymath}
To begin with, we evaluate the second term denoted as $F_{V2}$, as follows. The term inside the $\log$ can be evaluated as

\[
\begin{split}
  \pmb\Delta &= \frac{\Pr(\D| \cf, \tcf, \Z)\Pr(\cf|\bxi)\Pr(\tcf)\Pr(\bxi)}{q(\cf, \tcf, \bxi| \Z)}\\
  &= \frac{\prod_{j,k}\left[\Pr(\y^j|\tf^j, \ff_k)\right]^{z_{jk}} \prod_k\Pr(\ff_k|\bta_k)\prod_j\Pr(\tf^j)\prod_k\Pr(\bta_k)}
  {\prod_{j,k}\left[\Pr(\tf^j|\ff_k, \y^j)\right]^{z_{jk}}\prod_k\Pr(\ff_k|\bta_k)\phi(\bta_k)} \\
  &= \prod_{j=1}^m\prod_{k=1}^K\left[\frac{\Pr(\y^j|\ff_k, \tf^j)\Pr(\tf^j)}{\Pr(\tf^j|\ff_k, \y^j)}\right]^{z_{jk}}\times \prod_{k=1}^K\frac{\Pr(\bta_k)}{\phi(\bta_k)}. \end{split}
\]

Thus, we can write $F_{V2}$ as
\[
\begin{split}
F_{V2} &= \int q(\Z)q(\cf, \tcf, \bxi|\Z)(\log\pmb\Delta)\ d\cf d\tcf d\bxi d\Z\\
&=\int q(\Z) \left[\int\prod_{k=1}^K\phi(\bta_k)\left\{\log
G(\Z, \bxi, \D) +
\sum_{k=1}^K\log\left[\frac{\Pr(\bta_k)}{\phi(\bta_k)}\right]\right\}d\bxi\right]d\Z,
\end{split}
\]
where
\[
\log G(\Z, \bxi, \D) =
\int\Pr(\tcf|\cf, \Z)\Pr(\cf|\bxi)\log\left[\prod_{j=1}^M\prod_{k=1}^K\left[\frac{\Pr(\y^j|\ff_k, \tf^j)\Pr(\tf^j)}{\Pr(\tf^j|\ff_k, \y^j)}\right]^{z_{jk}}\right]d\cf
d\tcf.
\]
We show below that $\log G(\Z, \bxi, \D)$ can be decomposed
as
\[
\log G(\Z, \bxi, \D) = \sum_{j=1}^M\sum_{k=1}^Kz_{jk}\log
G(\bta_k, \y^j),
\]
where
\begin{align}
\label{eqn:indG}
  \log G(\bta_k, \y^j) = \log\left[\mathcal{N}(\y^j|\boldsymbol{\alpha}^k_j, \widehat{\fK}_{jj})\right] - \frac{1}{2}\textbf{Tr}\left[(\fK_{jj}^k - \fQ^k_{jj})\widehat{\fK}_{jj}^{-1}\right],
\end{align}
 where $\boldsymbol{\alpha}^k_j=\fK_{jk}\fK_{kk}^{-1}\bta_k$ and $\fQ_{jj}^k
 = \fK_{jk}\fK_{kk}^{-1}\fK_{kj}$. Consequently, the variational lower bound is

\[
\begin{split}
F_V  &= \int q(\Z)q(\pi)\log\left[\frac{\Pr(\bpi)\Pr(\Z|\bpi)}{q(\Z)q(\bpi)}\right]d\bpi d\Z \\
      &\quad +
      \int q(\Z) \left[\int\prod_{k=1}^K\phi(\bta_k)\left\{\log G(\Z, \bxi, \D) + \sum_{k=1}^K\log\left[\frac{\Pr(\bta_k)}{\phi(\bta_k)}\right]\right\}d\bxi\right]d\Z
\end{split}
\]

To optimize the parameters we use the variational EM algorithm.

\begin{itemize}
  \item In the \textbf{Variational E-Step}, we estimate $q^*(\Z), q^*(\bpi)$ and $\{\phi^*(\bta_k)\}$.

To get the variational distribution $q^*(\Z)$, we take derivative of $F_V$
w.r.t. $q(\Z)$ and set it to 0. This yields
  \[
  \begin{split}
    \log q^*(\Z) &= \int q(\bpi)\log(\Pr(\Z|\bpi))d\bpi + \int\prod_{k=1}^K\phi(\bta_k)\log G(\Z, \bxi, \D)d\bxi  \\
    &= \sum_{j=1}^M\sum_{k=1}^Kz_{jk}\left[\bbbe_{q(\bpi)}[\log\pi_k] + \bbbe_{\phi(\bta_k)}[\log G(\bta_k, \y^j)]\right] + \textbf{const}
  \end{split}
  \]
from which we can derive
 \[
 \begin{split}
   q^*(\Z) &= \prod_{j=1}^M\prod_{k=1}^K r_{jk}^{z_{jk}}, \qquad r_{jk} = \frac{\rho_{jk}}{\sum_{k=1}^K\rho_{jk}}\\
   \log\rho_{jk} &= \bbbe_{q(\bpi)}[\log\pi_k] + \bbbe_{\phi(\bta_k)}[\log G(\bta_k, \y^j)],
 \end{split}
 \]
 where $\bbbe_{q(\bpi)}[\log\pi_k] = \psi(\alpha_k) - \psi(\sum_k\alpha_k)$
 where $\psi$ is the digamma function, $\alpha_k$ is defined below in
 (\ref{eqn:alphak}), and $\bbbe_{\phi(\bta_k)}[\log G(\bta_k, \y^j)]$ is
 given below in (\ref{eqn:rjk2}).

For the variational distribution of $q^*(\bpi)$ the derivative yields
 \[
 \begin{split}
   \log q^*(\bpi) &= \log\Pr(\bpi) + \int q(\Z)\log(\Pr(\Z|\bpi))d\bpi + \textbf{const}\\
   &= (\alpha_0 - 1)\sum_{k=1}^K\log(\pi_k) + \sum_{j=1}^M\sum_{k=1}^K\bbbe[z_{jk}]\log\pi_k + \textbf{const}
 \end{split}
 \]
 and taking the exponential of both sides, we have
 \[
   q^*(\bpi) = \textbf{Dir}(\bpi|\pmb\alpha)
 \]
 where
 \begin{equation}
 \label{eqn:alphak}
 \alpha_k = \alpha_0 + N_k, \quad N_k = \sum_{j = 1}^Kr_{jk}.
 \end{equation}

The final step is to get the variational distribution of $\phi^*(\bta_k),
k=1,\cdots, K$. Notice that only $F_{V2}$ is a function of $\phi(\bta_k)$. We
can rewrite this portion as
 \begin{align}
 \label{eqn:phi}
   &\int\prod_{k=1}^K\phi(\bta_k) \left(\left\{\int q(\Z) \sum_{j=1}^M\sum_{k=1}^Kz_{jk}\left[\log G(\bta_k, \y^j)\right]d\Z\right\} + \sum_{k=1}^K\log\left[\frac{\Pr(\bta_k)}{\phi(\bta_k)}\right]\right)d\bxi\nonumber\\
   &\quad = \int\prod_{k=1}^K\phi(\bta_k) \left(\sum_{j=1}^M\sum_{k=1}^K\bbbe_{q(\Z)}[z_{jk}]\left[\log G(\bta_k, \y^j)\right] + \sum_{k=1}^K\log\left[\frac{\Pr(\bta_k)}{\phi(\bta_k)}\right]\right)d\bxi\nonumber\\
   &\quad = \sum_{k=1}^K\int\phi(\bta_k)\left\{\left[\sum_{j=1}^M\bbbe_{q(\Z)}[z_{jk}]\log G(\bta_k, \y^j)\right] + \log\left[\frac{\Pr(\bta_k)}{\phi(\bta_k)}\right]\right\}d\bta_k.
 \end{align}
 Thus, our task reduces to find $\phi^*(\bta_k)$ separately. Taking the derivative of (\ref{eqn:phi}) w.r.t. $\phi(\bta_k)$ and setting it to be zero, we have
 \[
   \log\phi^*(\bta_k) = \sum_{j=1}^M\bbbe_{q(\Z)}[z_{jk}]\log G(\bta_k, \y^j) + \log\Pr(\bta_k) + \textbf{const.}
 \]
Using (\ref{eqn:indG}) and the fact that second term in (\ref{eqn:indG}) is not a function of $\bta_k$, we obtain
 \begin{equation}
 \label{eqn:phistar}
 \phi^*(\bta_k) \propto \prod_{j=1}^M\left[\mathcal{N}(\y^{j}|\boldsymbol{\alpha}^k_j,
  \widehat{\fK}_{jj}^k)\right]^{\bbbe_{q(\Z)}[z_{jk}]}\Pr(\bta_k).
 \end{equation}
Thus, we have
\[
\phi^*(\bta_k) \propto \exp\left\{-\frac{1}{2}(\bta^k)^T\left(\fK_{kk}^{-1} \mathbf{\Phi}\fK_{kk}^{-1}\right)\bta^k +  (\bta_k)^T\left(\fK_{kk}^{-1}\sum_{j=1}^M\bbbe_{q(\Z)}[z_{jk}]\fK_{kj}[\widehat{\fK}_{jj}]^{-1}\y_j\right)\right\},
\]
where
 \[
 \mathbf{\Phi} = \fK_{kk} +
\sum_{j=1}^M\bbbe_{q(\Z)}[z_{jk}]\fK_{kj}[\widehat{\fK}_{jj}]^{-1}\fK_{jk}.
\]
Completing the square yields the Gaussian distribution
 \[
 \phi^*(\bta_k) = \mathcal{N}(\pmb\mu_k, \pmb\Sigma_k), \]
 where
 \begin{equation}
 \label{eqn:vardis}
 \pmb\mu_k = \fK_{kk}\mathbf{\Phi}^{-1}\sum_{j=1}^M\bbbe_{q(\Z)}[z_{jk}]\fK_{kj}[\widehat{\fK}_{jj}]^{-1}\y_j, \quad \pmb\Sigma_k = \fK_{kk}\mathbf{\Phi}^{-1}\fK_{kk}.
 \end{equation}
\item In the \textbf{Variational M-Step}, based on the previous estimated
variational distribution, we wish to find hyperparameters that maximize the
variational lower bound $F_V$. The terms that depend on the hyperparameters
and the inducing variables $\{\mathcal{X}^k_m\}$ are given in
(\ref{eqn:phi}). Therefore, using (\ref{eqn:indG}) again, we have
\[
\begin{split}
  F_V(\mathcal{X}_k,\boldsymbol\theta) &= \sum_{k=1}^K\int\phi^*(\bta_k)\left\{\left[\sum_{j=1}^Mr_{jk}\log G(\bta_k, \y^j)\right] + \log\left[\frac{\Pr(\bta_k)}{\phi^*(\bta_k)}\right]\right\}d\bta_k\\
  &= \sum_{k=1}^K\int\phi^*(\bta_k)\left\{\log\left[\sum_{j=1}^Mr_{jk}\mathcal{N}(\y^j|\boldsymbol{\alpha}^k_j, \widehat{\fK}_{jj})\right] + \log\left[\frac{\Pr(\bta_k)}{\phi^*(\bta_k)}\right]\right\}d\bta_k \\
  &\qquad\qquad - \frac{1}{2}\sum_{k=1}^K\sum_{j=1}^mr_{jk}\textbf{Tr}\left[(\fK_{jj}^k - \fQ_{jj}^k)\widehat{\fK}_{jj}^{-1}\right]\\
  &= \sum_{k=1}^K\int\phi^*(\bta_k)\left\{\log\left[\frac{\prod_{j=1}^M\left[\mathcal{N}(\y^{j}|\boldsymbol{\alpha}^k_j, \widehat{\fK}_{jj})\right]^{r_{jk}}\Pr(\bta_k)}{\phi^*(\bta_k)}\right]\right\}
  d\bta_k \\
  &\qquad\qquad- \frac{1}{2}\sum_{k=1}^K\sum_{j=1}^mr_{jk}\textbf{Tr}\left[(\fK_{jj}^k - \fQ_{jj}^k)\widehat{\fK}_{jj}^{-1}\right]
\end{split}
\]
From (\ref{eqn:phistar}), we know that the term inside the log
is constant, and therefore, extracting the log from the integral and
cancelling the $\phi*$ terms we see that the $k$'th element of first term is
equal to the logarithm of
\begin{equation}
 \int\prod_{j=1}^M\left[\mathcal{N}(\y^{j}|\boldsymbol{\alpha}^k_j, \widehat{\fK}_{jj})\right]^{r_{jk}}\Pr(\bta_k)d\bta_k.
\end{equation}
We next show how this multivariate integral can be evaluated. First consider
\[
\begin{split}
&\left[\mathcal{N}(\y^{j}|\boldsymbol{\alpha}^k_j, \widehat{\fK}_{jj})\right]^{r_{jk}} \\
&\quad =
\left((2\pi)^{-\frac{N_j}{2}}|\widehat{\fK}_{jj}|^{-\frac{1}{2}}\right)^{r_{jk}}\exp\left\{-\frac{r_{jk}}{2}(\y^{j} - \boldsymbol{\alpha}^k_j)^T[\widehat{\fK}_{jj}]^{-1}(\y^{j} - \boldsymbol{\alpha}^k_j)\right\}\\
&\quad =\left((2\pi)^{-\frac{N_j}{2}}|\widehat{\fK}_{jj}|^{-\frac{1}{2}}\right)^{r_{jk}}\exp\left\{-\frac{1}{2}(\y^{j} - \boldsymbol{\alpha}^k_j)^T\left[r_{jk}^{-1}\widehat{\fK}_{jj}\right]^{-1}(\y^{j} - \boldsymbol{\alpha}^k_j)\right\}\\
&\quad = \frac{\left((2\pi)^{-\frac{N_j}{2}}|\widehat{\fK}_{jj}|^{-\frac{1}{2}}\right)^{r_{jk}}}
{(2\pi)^{-\frac{N_j}{2}}|r_{jk}^{-1}\widehat{\fK}_{jj}|^{-\frac{1}{2}}}\cdot (2\pi)^{-\frac{N_j}{2}}|r_{jk}^{-1}\widehat{\fK}_{jj}|^{-\frac{1}{2}}\exp\left\{-\frac{1}{2}(\y^{j} - \boldsymbol{\alpha}^k_j)^T\left[r_{jk}^{-1}\widehat{\fK}_{jj}\right]^{-1}(\y^{j} - \boldsymbol{\alpha}^k_j)\right\}\\
&\quad = A_{jk}\mathcal{N}(\y^{j}|\boldsymbol{\alpha}^k_j, r_{jk}^{-1}\widehat{\fK}_{jj}^k),
\end{split}
\]
where $A_{jk} = (r_{jk})^{\frac{N_j}{2}}(2\pi)^{\frac{N_j(1-r_{jk})}{2}}|\widehat{\fK}_{jj}|^{\frac{1-r_{jk}}{2}}$. Thus, we have
\[
\prod_{j=1}^M\left[\mathcal{N}(\y^{j}|\boldsymbol{\alpha}^k_j,
  \widehat{\fK}_{jj})\right]^{r_{jk}} = \left[\prod_{j=1}^M A_{jk}\right]\prod_{j=1}^M\mathcal{N}(\y^{j}|\boldsymbol{\alpha}^k_j, r_{jk}^{-1}\widehat{\fK}_{jj}).
\]
The first part is not a function of $\bta_k$, for the integration we are only interested in the second part. Since $\breve{\y}$ is the concatenation of all $\y^j$'s, we can write
\begin{equation}
\label{eqn:fvreduce}
\prod_{j=1}^M\mathcal{N}(\y^{j}|\boldsymbol{\alpha}^k_j, r_{jk}^{-1}\widehat{\fK}_{jj}) = \mathcal{N}(\breve{\y}|\mathbf\Lambda_k\fK_{kk}^{-1}\bta_k, \widehat{\fK}^k),
\end{equation}
where
\[
\mathbf\Lambda_k = \begin{pmatrix}
  \fK_{1k}\\
  \fK_{2k}\\
  \vdots\\
  \fK_{Mk}
\end{pmatrix} \qquad \text{and}\qquad
\widehat{\fK}^k = \bigoplus_{j=1}^M r_{jk}^{-1}\widehat{\fK}_{jj}^k = \begin{pmatrix}
  r_{1k}^{-1}\widehat{\fK}_{11} & & &\\
  & r_{2k}^{-1}\widehat{\fK}_{22} & & \\
  & & \ddots & \\
  & & & r_{Mk}^{-1}\widehat{\fK}_{MM}
\end{pmatrix}.
\]
Therefore, the integral can the written as the following marginal distribution of $\Pr(\breve{\y}|k)$,
\begin{equation}
\label{eqn:fvmarginal}
 \int\prod_{j=1}^M\mathcal{N}(\y^{j}|\boldsymbol{\alpha}^k_j, r_{jk}^{-1}\widehat{\fK}_{jj})\Pr(\bta_k)d\bta_k = \int\mathcal{N}(\breve{\y}|\mathbf\Lambda_k\fK_{kk}^{-1}\bta_k, \widehat{\fK}^k)\Pr(\bta_k)d\bta_k = \Pr(\breve{\y}|k).
\end{equation}

Using the fact that $\Pr(\bta_k) = \mathcal{N}(\mathbf{0}, \fK_{kk})$ and
observing that (\ref{eqn:fvreduce}) is a conditional Gaussian, we have
\[
\Pr(\breve{\y}|k) = \mathcal{N}(\mathbf{0}, \mathbf\Lambda_k\fK_{kk}^{-1}\mathbf\Lambda_k^T + \widehat{\fK}^k).
\]
Using this form and the portion of $A_{jk}$ that depends on the parameters we get
\begin{align}
\label{eqn:vlbfinal}
F_V(\mathcal{X},\boldsymbol\theta) &= \sum_{k=1}^K\log\Pr(\breve{\y}|k) + \sum_{k=1}^K\sum_{j=1}^M\frac{1-r_{jk}}{2}\log|\widehat{\fK}_{jj}| - \frac{1}{2}\sum_{k=1}^K\sum_{j=1}^Mr_{jk}\textbf{Tr}\left[(\fK_{jj}^k - \fQ_{jj}^k)\widehat{\fK}_{jj}^{-1}\right]\nonumber\\
&= \sum_{k=1}^K\log\Pr(\breve{\y}|k) + \frac{K - 1}{2}\sum_{j=1}^M\log|\widehat{\fK}_{jj}| - \frac{1}{2}\sum_{k=1}^K\sum_{j=1}^Mr_{jk}\textbf{Tr}\left[(\fK_{jj}^k - \fQ_{jj}^k)\widehat{\fK}_{jj}^{-1}\right]
\end{align}
This extends the bound for the single center $K=1$ case given in~(\ref{eqn:lb}). Furthermore, following the same line as the previous derivation, the direct inference for the full model can be obtained where $\eta_k$ is substituted with $\ff_k$ and the variational lower bound becomes
\[
F_V(\mathcal{X},\boldsymbol\theta) = \sum_{k=1}^K\log\mathcal{N}(\breve{y}| \mathbf{0}, \fK_{kk} + \widehat{\fK}^k) + \frac{K - 1}{2}\sum_{j=1}^M\log|\widehat{\fK}_{jj}|.
\]

We have explicitly written the parameters that can be chosen
to further optimize the lower bound (\ref{eqn:vlbfinal}), namely the support inputs
$\{\mathcal{X}^k_m\}$, and the hyper-parameters
$\boldsymbol\theta$ which are composed of
$\{\boldsymbol\theta_k\}$ and
$\{\tilde{\pmb\theta}\}$ in ${K_k}$ and $\widetilde{{K}}$
respectively.

By calculating derivatives of (\ref{eqn:vlbfinal}) we
can optimize the lower bound using a gradient based method. It is easy to see
that the complexity for calculating the derivative of the second and third
terms of (\ref{eqn:vlbfinal}) is $\mathcal{O}(N)$. Thus, the key
computational issue of deriving a gradient descent algorithm involves
computing the derivative of $\log\Pr(\breve{\y}|k)$.
We first show how to calculate the inverse of
the
$N\times N$ matrix $\mathbf\Upsilon =
\mathbf\Lambda_k\fK_{kk}^{-1}\mathbf\Lambda_k^T + \widehat{\fK}^k$.
Using the matrix inversion lemma (the Woodbury identity),
we have
\[
(\mathbf\Lambda_k\fK_{kk}^{-1}\mathbf\Lambda_k^T + \widehat{\fK}^k)^{-1} = [\widehat{\fK}^k]^{-1} - [\widehat{\fK}^k]^{-1}\mathbf\Lambda_k \left( \fK_{kk} + \mathbf\Lambda_k^T[\widehat{\fK}^k]^{-1}\mathbf\Lambda_k\right)^{-1}\mathbf\Lambda_k^T[\widehat{\fK}^k]^{-1}.
\]
Since $\widehat{\fK}^k$ is a block-diagonal matrix, its inverse
can be calculated in $\sum_j\mathcal{O}(N_j^3)$.
Now,
$\fK_{kk} +
\mathbf\Lambda_k^T[\widehat{\fK}^k]^{-1}\mathbf\Lambda_k$ is an $m_k\times
m_k$ matrix where $m_k$ is the number of inducing variables for the $k$-th
mean effect. Therefore the computation of (\ref{eqn:vlbfinal}) can be done in
$\mathcal{O}(m_k^3 + \sum_jN_j^3 + Nm_k^2)$.
Next,
consider calculating the
derivative of the first term. We have
\[
\frac{\partial \Pr(\breve{\y}|k)}{\partial\theta_j} = \frac{1}{2}\breve{\y}^T\mathbf\Upsilon^{-1}\frac{\partial{\mathbf\Upsilon}}{\partial\theta_j}\mathbf\Upsilon ^{-1}\breve{\y} - \frac{1}{2}\textbf{Tr}(\mathbf\Upsilon^{-1}\frac{\partial{\mathbf\Upsilon}}{\partial\theta_j}),
\]
where, by the chain rule, we have
\[
\frac{\partial{\mathbf\Upsilon}}{\partial\theta_j} = \frac{\partial\mathbf\Lambda_k}{\partial\theta_j}\fK_{kk}^{-1}\mathbf\Lambda_k^T - \mathbf\Lambda_k\fK_{kk}^{-1}\frac{\partial\fK_{kk}}{\partial\theta_j}\fK_{kk}^{-1}\mathbf\Lambda_k^T +\mathbf\Lambda_k\fK_{kk}^{-1}\frac{\partial\mathbf\Lambda_k^T}{\partial\theta_j} + \frac{\partial\widehat{\fK}^k}{\partial\theta_j}.
\]
Therefore, pre-calculating
$\breve{\y}^T\mathbf\Upsilon^{-1}$
and sequencing the other matrix operations
from left to right
the gradient calculation for each hyperparameter can be calculated in
$\mathcal{O}(Nm_k^2)$. In our implementation, we use stochastic coordinate descent, where at each iteration, one coordinate (parameter) is chosen at random and we perform gradient descent on that coordinate.
\end{itemize}
\subsection{Evaluating $\log G(\Z, \bxi, \D)$}
In this section, we develop the expression for $\log G(\Z, \bxi, \D)$.
\begin{equation}
  \label{eqn:logG}
  \begin{split}
    &\log G(\Z, \bxi, \D) \\
    &\ = \int \prod_{l=1}^M\prod_{p=1}^K\left[\Pr(\tf^l|\ff_p, \y^l)\right]^{z_{lp}}\prod_{v=1}^K\Pr(\ff_v|\bta_v)
    \times\sum_{j=1}^M\sum_{k=1}^K z_{jk}\log\left[\frac{\Pr(\y^j|\ff_k, \tf^j)\Pr(\tf^j)}{\Pr(\tf^j|\ff_k, \y^j)}\right]d\cf d\tcf\\
    &\ = \sum_{j=1}^M\sum_{k=1}^K z_{jk}\left[\int \left(\prod_{v=1}^K\Pr(\ff_v|\bta_v)\right) \times \Pr(\tf^j|\ff_k, \y^j)\times \log\left[\frac{\Pr(\y^j|\ff_k, \tf^j)\Pr(\tf^j)}{\Pr(\tf^j|\ff_k, \y^j)}\right]d\cf
    d\tf^j\right]\\
    &\ = \sum_{j=1}^M\sum_{k=1}^K z_{jk}\left[\int  \Pr(\tf^j|\ff_k, \y^j)\times \left[\int \Pr(\ff_k|\bta_k)\prod_{v=1,v\neq k}^K\Pr(\ff_v|\bta_v)d\cf_{-k}\right] \times \log\left[\frac{\Pr(\y^j|\ff_k, \tf^j)\Pr(\tf^j)}{\Pr(\tf^j|\ff_k, \y^j)}\right]d\ff_k\tf^j\right]\\
    &\ = \sum_{j=1}^M\sum_{k=1}^K z_{jk}\left[\int \Pr(\tf^j|\ff_k, \y^j)\Pr(\ff_k|\bta_k)\times \log\left[\frac{\Pr(\y^j|\ff_k, \tf^j)\Pr(\tf^j)}{\Pr(\tf^j|\ff_k, \y^j)}\right]d\ff_k
    d\tf^j\right]\\
    &\ = \sum_{j=1}^m\sum_{k=1}^Kz_{jk}\log G(\bta_k, \y_j)
  \end{split}
\end{equation}
where the second line holds because in the sum indexed by $j$ and $k$ all
the product measures
\[\prod_{l=1, l\neq j}^M\prod_{p=1}^K\left[\Pr(\tf^l|\ff_p, \y^l)\right]^{z_{lp}},\]  are
integrated to 1, leaving only the $\Pr(\tf^j|\ff_k, \y^j)$. Our next step is to evaluate $\log G(\bta_k, \y_j)$, we have
\begin{align}
\label{eqn:hatG}
    &\int \Pr(\tf^j|\ff_k, \y^j)\Pr(\ff_k|\bta_k)\times \log\left[\frac{\Pr(\y^j|\ff_k, \tf^j)\Pr(\tf^j)}{\Pr(\tf^j|\ff_k, \y^j)}\right]d\ff_k
    d\tf^j\nonumber\\
    &\ =\int\Pr(\tf^j|\ff_k, \y^j)\Pr(\ff_k|\bta_k)\times\log\left[\Pr(\y^j|\ff_k, \tf^j)\Pr(\tf^j)\cdot\frac{\Pr(\y^j|\ff_k)}{\Pr(\y^j|\ff_k, \tf^j)\Pr(\tf^j|\ff_k)}\right]d\ff_k d\tf^j\nonumber\\
    &\ = \int\Pr(\ff_k|\bta_k)\log\left[\Pr(\y^j|\ff_k)\right]d\ff_k = \int \Pr(\f^j|\bta_k)\log\left[\Pr(\y^j|\f^j)\right]d\f^j
\end{align}
where the last line holds because of the independence between $\tf^j$
and $\ff_k$. We next show how this expectation can be evaluated.
This is more complex than the single-task case because of the coupling of the fixed-effect and the random effect.

Recall that $
  \Pr(\f^j|\bta_k) = \mathcal{N}(\f^j|\fK_{jk}\fK_{kk}^{-1}\bta_k, \fK_{jj}^k-\fK_{jk}{\fK}_{kk}^{-1}{\fK}_{kj})
$.
Denote $\widehat{\fK}_{jj}^{-1}=\fL^\T\fL$
where $\fL$ can be chosen as its Cholesky factor, we have
\begin{displaymath}
\begin{split}
   \log\left[\Pr(\y^j|\f^j)\right]
   &= -\frac{1}{2}(\fL\y^j - \fL\f^j)^T(\fL\y^j - \fL\f^j)+ \log\left[(2\pi)^{-\frac{N_j}{2}}\right] + \log\left[|\widehat{\fK}_{jj}|^{-\frac{1}{2}}\right].
\end{split}
\end{displaymath}
Notice that $\Pr(\fL\f^j|\bta_k) =
\mathcal{N}(\fL\fK_{jk}\fK_{kk}^{-1}\bta_k, \fL(\fK^k_{jj} -
\fQ^k_{jj})\fL^\T)$ where $\fQ^k_{jj} = \fK_{jk}\fK_{kk}^{-1}\fK_{kj}$.
Thus,
\[
\begin{split}
  &\log G(\bta_k, \y^j) = \bbbe_{[\f^j|\bta_k]}\log\left[\Pr(\y^j|\f^j)\right]\\
  &\ =-\frac{1}{2}\bigg\|\fL\y^j - \fL\fK_{jk}\fK_{kk}^{-1}\bta_k\bigg\|^2 -\frac{1}{2}\textbf{Tr}(\fL(\fK_{jj}^k - \fQ^k_{jj})\fL^T) + \log\left[(2\pi)^{-\frac{N_j}{2}}\right] + \log\left[|\widehat{\fK}_{jj}|^{-\frac{1}{2}}\right]\\
  &\ = \log\left[\mathcal{N}(\y^j|\boldsymbol{\alpha}_j, \widehat{\fK}_{jj})\right] - \frac{1}{2}\textbf{Tr}\left[(\fK_{jj}^k - \fQ^k_{jj})\widehat{\fK}_{jj}^{-1}\right]
\end{split}
\]
where $\boldsymbol{\alpha}^k_j=\fK_{jk}\fK_{kk}^{-1}\bta_k$. Finally, we have
\begin{equation}
\label{eqn:Gfm}
\begin{split}
  \log G(\bxi, \D)
  &= \sum_{j=1}^m\sum_{k=1}^Kz_{jk}\Bigg[\log\left[\mathcal{N}(\y^{j}|\boldsymbol{\alpha}^k_j,
  \widehat{\fK}_{jj})\right]
  - \frac{1}{2}\textbf{Tr}\left[(\fK^k_{jj} - \fQ^k_{jj})[\widehat{\fK}_{jj}]^{-1}\right]\Bigg].
\end{split}
\end{equation}
Furthermore, marginalization out $\bta_k$, we have
\begin{equation}
  \label{eqn:rjk2}
  \bbbe_{\phi^*(\bta_k)}\log G(\bta_k, \y^j) = \log\left[\mathcal{N}(\y^j|\boldsymbol{\mu}_j^k, \widehat{\fK}_{jj})\right] - \frac{1}{2}\textbf{Tr}\left[\fK_{jk}\fK_{kk}^{-1}(\pmb\Sigma_k - \fK_{kk})\fK_{kk}^{-1}\fK_{jk}\widehat{\fK}_{jj}^{-1}\right]
\end{equation}

\subsection{Prediction Using the Sparse Model}
The proposed sparse model can be used for two types of problems. Prediction for existing tasks and prediction for a newly added task. We start with deriving the predictive distribution for existing tasks. Given any task $j$, our goal is to calculate the predictive
distribution $\Pr(f^j(\x^*)|\D)$ at
 new input point $\x^*$, which can be written as

{{\begin{align}
\label{eqn:bapred}
\sum_{k=1}^K\Pr(f^j(\x^*)|z_{jk}=1, \D)\Pr(z_{jk}=1|\D)= \sum_{k=1}^Kr_{jk}\Pr(f^j(\x^*)|z_{jk} = 1, \D).
\end{align}}}

That is, because $z_{jk}$ form a partition we can focus on calculating $\Pr(f^j(\x^*)|$ $z_{jk} = 1, \D)$ and then combine the results using the partial labels. Calculating (\ref{eqn:bapred}) is exactly the same as the predictive distribution in the non-grouped case, the derivation in Section 3.2 gives the details. The complexity of these computations is $\mathcal{O}(K(N_j^3 +
m^3))$ which is a significant improvement over $\mathcal{O}(KN^3)$ where $N=\sum_jN_j$. Instead of calculating the full Bayesian prediction, one can use \emph{Maximum A Posteriori} (MAP) by assigning the $j$-th task to the center $c$ such that $c = \text{argmax}\Pr(z_{jk}=1|\D)$. Preliminary experiments (not shown here) show that the full Bayesian approach gives better performance. Our experiment below is the results of Bayesian prediction.

Our model is also useful for making prediction for newly added tasks. Suppose we are given $\{\x^{M+1}, \y^{M+1}\}$ and we are interested in predicting $f^{M+1}(\x^*)$. We use the variational procedure to estimate its partial labels w.r.t. different centers $\Pr(z_{M+1,k}=1|\mathcal{D})$ and then (\ref{eqn:bapred}) can be applied for making the prediction. In the variational procedure we update the parameters for $Z_{M+1}$ but keep all other parameters fixed. Since each task has small number of samples, we expect this step to be computationally cheap.

\section{Related Work}
\label{sec:related}
Our work is related to~\citep{titsias2008variational}
particularly in terms of the form of the variational distribution of
the inducing variables. However, our model is much more
complex than the basic GP regression model. With the mixture model and an additional random effect per task, we must take into account the
coupling of the random effect and group specific fixed-effect functions. The
technical difficulty that the coupling introduces is addressed in
our paper, yielding a generalization that is consistent with the single-task solution.

The other related thread comes from the area of GP for multi-task
learning. \cite{bonilla2008multi}proposed a model that learns a shared covariance
matrix on features and a covariance matrix for tasks that explicitly models
the dependency between tasks. They also presented
techniques to speed up the inference by using Nystrom approximation of the
kernel matrix and incomplete Cholesky decomposition of the task correlation
matrix. Their model, which is known as the linear coregionalization model
(LCM) is subsumed by the framework of convolved multiple output Gaussian
process~\citep{alvarez2011computationally}.
The work of
\cite{alvarez2011computationally}
also derives sparse solutions which are extensions of different single task
sparse GP~\citep{Snelson06,quinonero2005unifying}. Our work differs from the
above models in that we allow a random effect for each individual task. As we
show in the experimental section, this is important in modeling various
applications. If the random effect is replaced with independent white noise,
then our model is similar to LCM. To see this, from (\ref{eqn:bapred}), we
recognize that the posterior GP is a convex combination of $K$ independent
GPs (mean effect). However, our model is capable of prediction for newly
added tasks while the models in ~\citep{bonilla2008multi}
and~\citep{alvarez2011computationally} cannot. Further, the proposed model can
naturally handle \textit{heterotopic} inputs, where different tasks do not
necessarily share the same inputs. In~\citep{bonilla2008multi}, each task is
required to have same number of samples so that one can use the property of
Kronecker product to derive the EM algorithm. 
\section{Experimental Evaluation}

Our implementation of the algorithm makes use of the
gpml package \citep{rasmussen2010gaussian} 
and extends it to implement the required functions.
For performance criteria we use the standardized mean square error (SMSE)
and the mean standardized log loss (MSLL) that are defined
in~\citep{rasmussen2005gaussian}. We compare the following methods. The first four methods use the same variational inference as described in Section 4. They differ in the form of the variational lower bound they choose to optimize.
\begin{enumerate}
\item
\textbf{Direct Inference}: use full samples as the support variables and optimize the marginal likelihood. When $K=1$, the marginal likelihood is described in Section~\ref{sec:sinfer} and the predictive distribution is~(\ref{eq:fullpredictive}).
\item
\textbf{Variational Sparse GP for MTL (MT-VAR)}: the proposed approach.
\item
\textbf{MTL Subset of Datapoints (MT-SD)}: a subset $\mathcal{X}^k_m$ of size $m_k$ is chosen
  uniformly from the input points from all tasks $\breve{\x}$ for each center.  The
  hyper-parameters are selected using $\mathcal{X}_m^k$ (the inducing variables are fixed in advance) and their
  corresponding observations by maximizing the variational lower bound. We call this MT-SD as a multi-task version of
  SD~\citep[see][chap. 8.3.2]{rasmussen2005gaussian}, because in the single center case, this method uses (\ref{eq:fullMlikelihood}) and~(\ref{eq:fullpredictive}) using the subset
  $\mathcal{X}_m, \mathcal{Y}_m$ and $\x^j, \y^j$ as the full sample (thus discarding other samples).
\item
\textbf{MTL Projected Process Approximation (MT-PP)}:  the variational lower bound of MT-PP is given by
  the first two terms of (\ref{eqn:vlbfinal}) ignoring the trace term, and therefore the optimization
  chooses different pseudo inputs and hyper-parameters. We call it MT-PP because in the single center case, it corresponds to a multi-task version of
  PP~\citep[see][chap. 8.3.3]{rasmussen2005gaussian}.
\item
\textbf{Convolved Multiple Output GP (MGP-FITC, MGP-PITC)}: the approaches proposed in~\citep{alvarez2011computationally}. For all experiments, we use code from ~\citep{alvarez2011computationally} with the following setting. The kernel type is set to be \texttt{gg}. The hyperparameters, parameters and the position of inducing variables are obtained via optimizing the marginal likelihood using a scaled conjugated gradient algorithm. The support variables are initialized as equally spaced points over the range of the inputs. We set the $R_q = 1$, which means that the latent functions share the same covariance function. Whenever possible, we set $Q$ which, roughly speaking, corresponds to the number of centers in our approach, to agree with the number of centers. The maximum number of iterations allowed in the optimization procedure is set to be 200. The number of support variables is controlled in the experiments as in our methods.
\end{enumerate}

Three datasets are used to demonstrate the empirical performance of the
proposed approach. The first synthetic dataset contains data sampled
according to our model.  The second dataset is also synthetic but it is
generated from differential equations describing glucose concentration in
biological experiments, a problem that has been previously used to evaluate
multi-task GP~\citep{pillonetto2010bayesian}. Finally, we apply the proposed
method on a real astrophysics dataset. For all experiments, the kernels for
different centers are assumed to be the same. The hyperparameter for the
Dirichlet distribution is set to be $\alpha_0 = 1/K$. Unless otherwise specified, the inducing variables
are initialized to be equally spaced points over the range of the inputs.
To initialize, tasks are randomly assigned into groups. We run the conjugate gradient
algorithm (\texttt{minimize.m}) on a small subset of tasks (100 tasks each
having 5 samples) to get the starting values of hyperparameters of the
$\widetilde{\cK}$ and $\cK$, and then follow with the full optimization as above.
Finally, we repeat the entire procedure 5 times
and choose the one that achieves best variational lower bound. The maximum
number of iterations for the stochastic coordinate descent is set to be 50
and the maximum number of iterations for the variational inference is set to
be 30. The entire experiment is repeated 10 times to obtain the average
performance and error bars.

\subsection{Synthetic data}

In the first experiment, we demonstrate the performance of our
algorithm on a regression task with artificial data. More precisely,
we generated 1000 single-center tasks where each $f^j(x) = \bar{f}(x) +
\tilde{f}^j(x)$ is generated on the interval $x\in[-10, 10]$. Each
task has 5 samples. The fixed-effect function is sampled from a GP
with covariance function \[ \textbf{Cov}[\bar{f}(t_1), \bar{f}(t_2)]
= e^{-(t_1-t_2)^2/2}.\] The individual effect $\tilde{f}^j$ is
sampled via a GP with the covariance function \[
\textbf{Cov}[\tilde{f}^j(t_1), \tilde{f}^j(t_2)] =
0.25e^{-(t_1-t_2)^2/2}.\] The noise level $\sigma^2$ is set to be
$0.1$. The sample points $\x^j$ for each task are sampled uniformly
in the interval $[-10, 10]$ and the 100 test samples are chosen equally spaced in the same interval. The fixed-effect curve is generated by
drawing a single realization from the distribution of $\bar{\f}$
while the $\{\f^j\}$ are sampled i.i.d. from their common prior.
We set the number of latent functions $Q=1$ for MGP.

\begin{figure}[t]
{
        \begin{minipage}{0.5\textwidth}
        \centering
        \includegraphics[width=\textwidth]{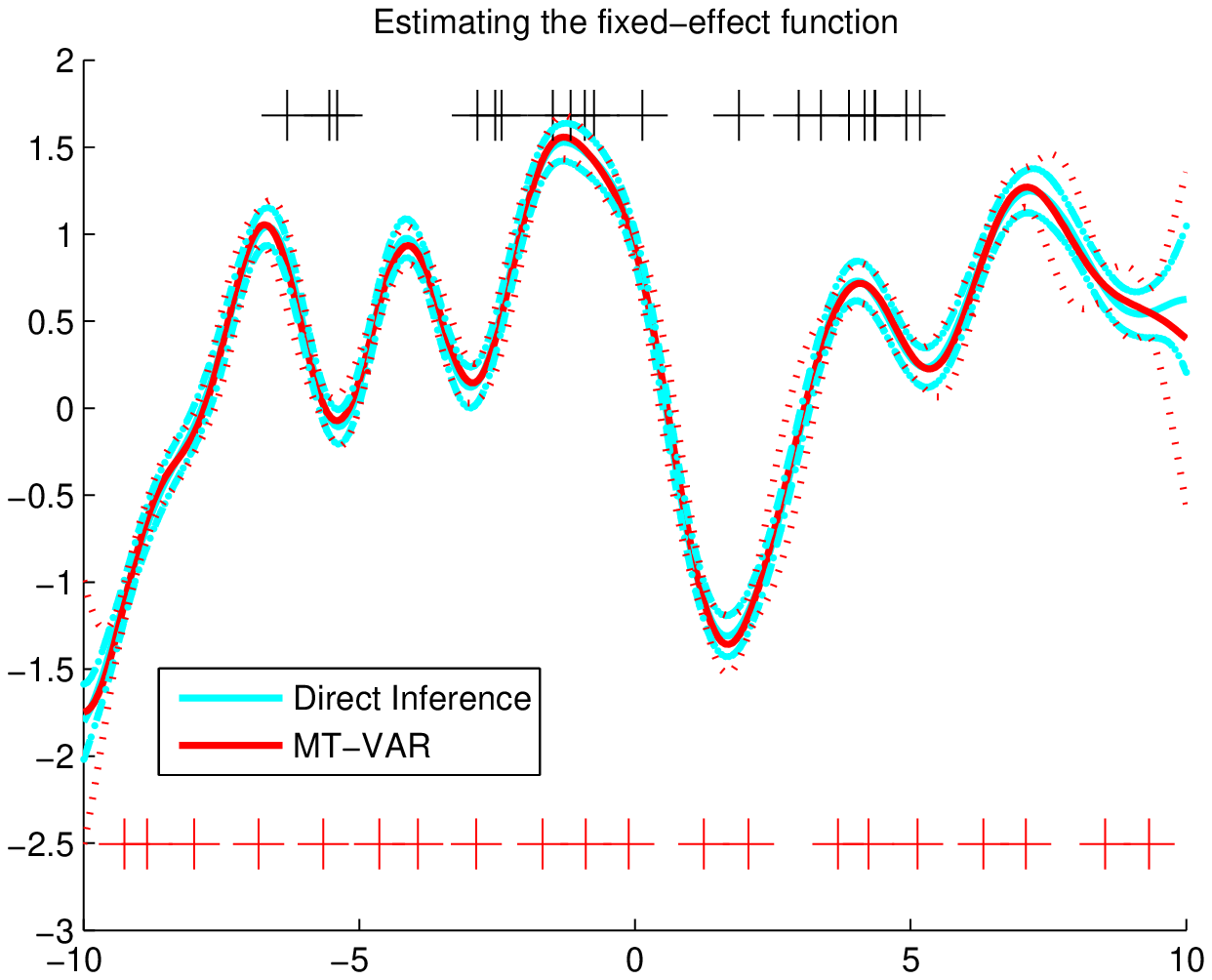}
        \end{minipage}
        \begin{minipage}{0.5\textwidth}
        \centering
        \includegraphics[width=\textwidth]{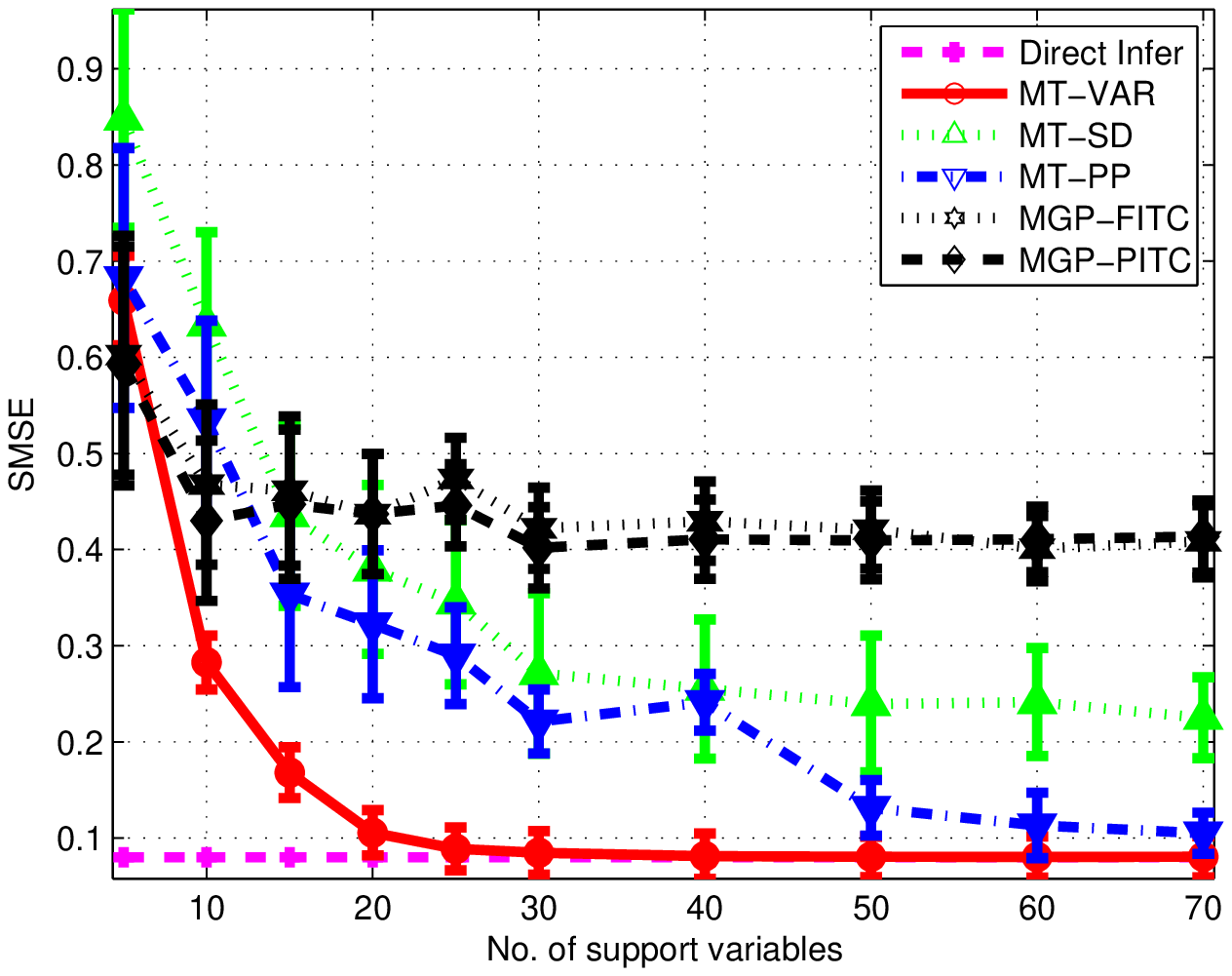}
        \end{minipage}\hfill
        \begin{minipage}{0.5\textwidth}
        \centering
        \includegraphics[width=\textwidth]{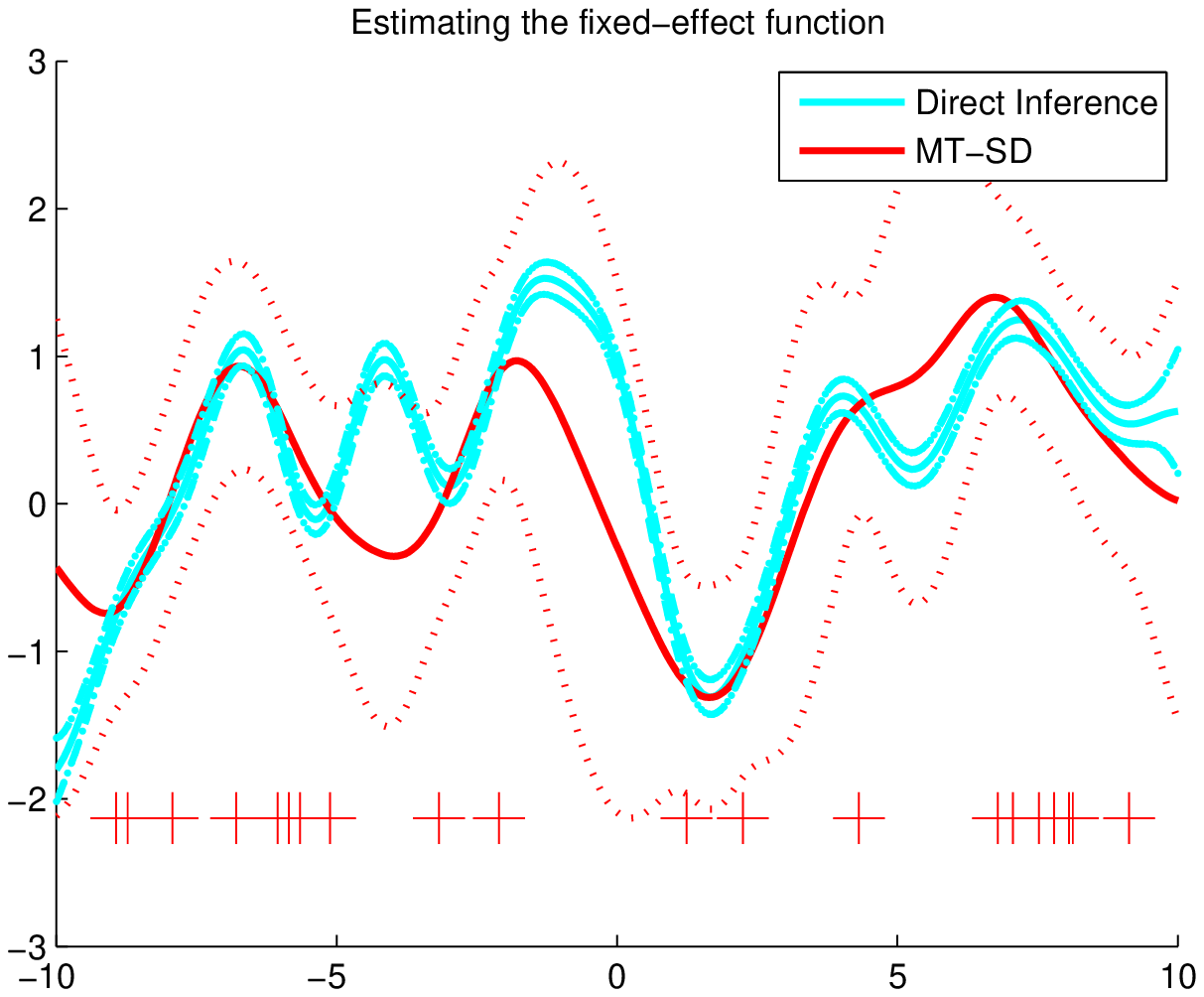}
        \end{minipage}
        \begin{minipage}{0.5\textwidth}
        \centering
        \includegraphics[width=\textwidth]{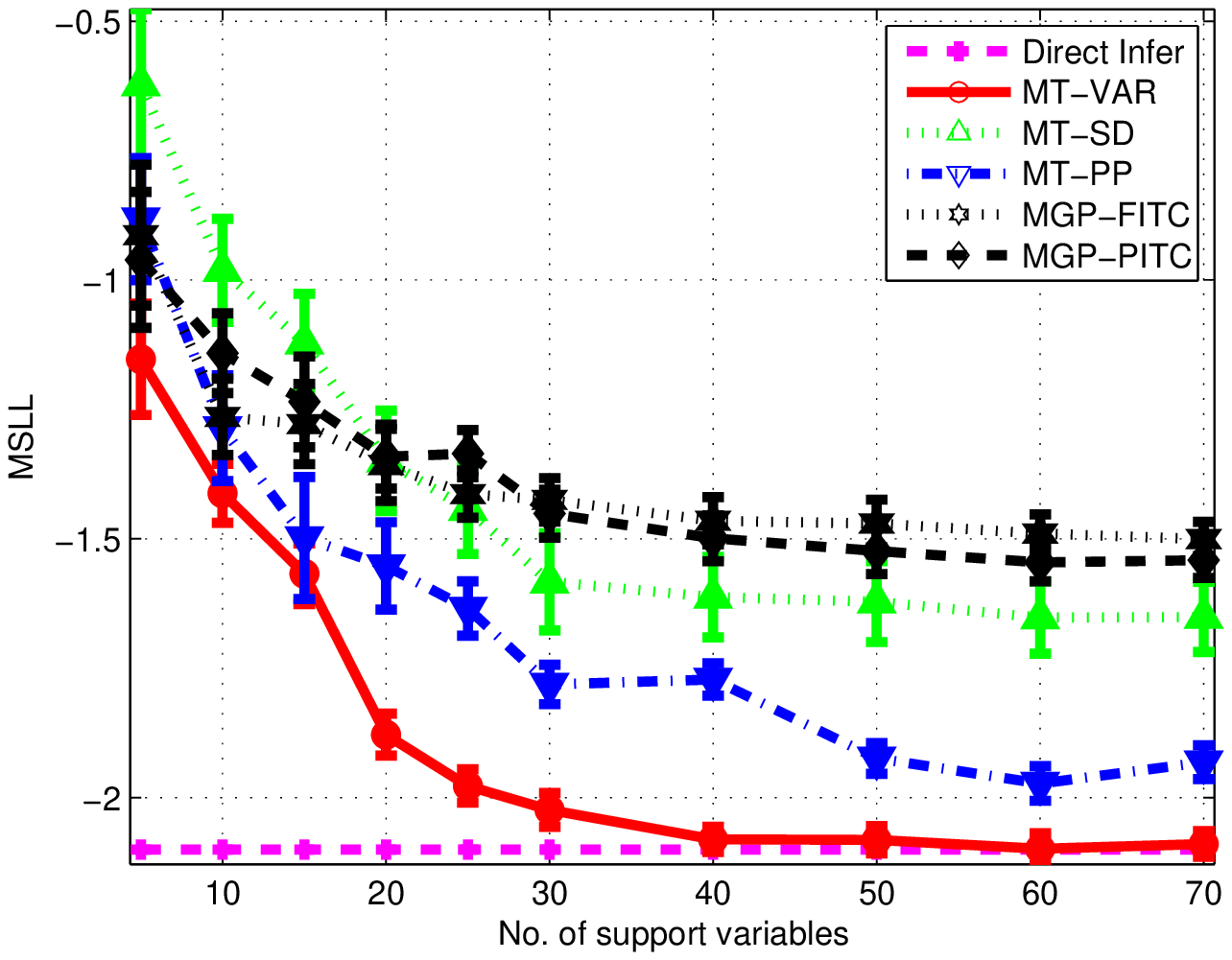}
        \end{minipage}\hfill
        \begin{minipage}{0.5\textwidth}
        \centering
        \includegraphics[width=\textwidth]{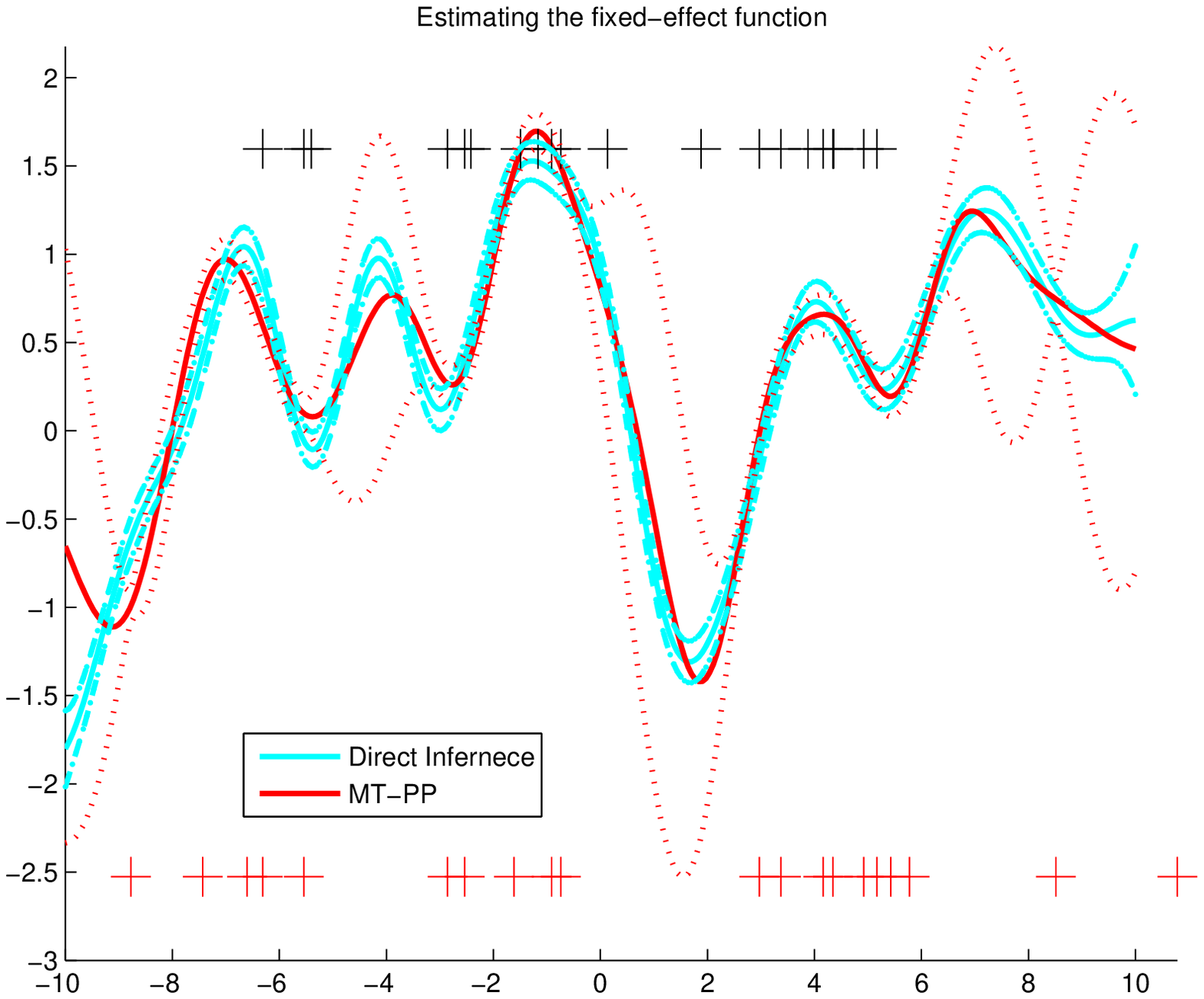}
        \end{minipage}
        \vspace{-0.3cm}
\caption{ Synthetic Data: Comparison between the proposed method and
other approaches. Left Column:
Predictive distribution for the fixed-effect. The solid line denotes
the predictive mean and the corresponding dotted line is the
predictive variance. The black crosses are the initial value of the inducing variables and the red ones are their values after learning process. Right Column: The average SMSE and MSLL for all the tasks.
}
\label{fig:synthetic}}
\end{figure}

The results are
shown in Fig.~(\ref{fig:synthetic}).
The top row shows qualitative results for one run using 20 support variables. We restrict the initial support variables to be in $[-7, 7]$ on purpose to show that the proposed method is capable of finding the optimal inducing variables.
It is clear that the predictive distribution of the proposed method
is much closer to the results of direct inference. The bottom row
gives quantitative results for SMSE and MSLL showing the same, as
well as showing that with 40 pseudo inputs the proposed method
recovers the performance of full inference. The MGP performs poorly on this dataset, indicating that it is not sufficient to capture the random effect. We also see a large computational advantage over MGP in this experiment. When the number of inducing variables is 20,  the training time for FITC (the time for constructing the sparse model plus the time for optimization) is 1515.19 sec. while the proposed approach is about 7 times faster (201.81 sec.)\footnote{The experiment was performed using \textsc{matlab} R2012a on an Intel Core Quo 6600 powered Windows 7 PC with 4GB memory.}.

\subsection{Simulated Glucose Data}
We evaluate our method to reconstruct the glucose profiles in an
intravenous glucose tolerance test
(IVGTT)~\citep{vicini2001iterative,denti2010ivgtt,pillonetto2010bayesian}
where \cite{pillonetto2010bayesian} developed an online multi-task
GP solution for the case where sample points are frequently shared
among tasks. This provides a more realistic test of our algorithm
because data is not generated explicitly by our model. More precisely, we apply the algorithm to reconstruct the
glucose profiles in an intravenous glucose tolerance test (IVGTT)
where blood samples are taken at irregular intervals of time,
following a single intravenous injection of glucose.
We generate the data using minimal models of glucose which is
commonly used to analyze glucose and insulin IVGTT
data~\citep{vicini2001iterative}, as follows~\citep{denti2010ivgtt}
\begin{equation}
\begin{split}
\dot{G}(t) &= -[S_G + X(t)]G(t) + S_G\cdot G_b + \delta(t)\cdot D/V\\
\dot{X}(t) &= -p_2\cdot X(t) + p_2\cdot S_I\cdot[I(t) - I_b]\\
G(0) &= G_b, \quad X(0) = 0
\end{split}
\end{equation}
where $D$ denotes the glucose dose, $G(t)$ is plasma glucose
concentration and $I(t)$ is the plasma insulin concentration which
is assumed to be known. $G_b$ and $I_b$ are the glucose and insulin
base values. $X(t)$ is the insulin action and $\delta(t)$ is the
Dirac delta function. $S_G, S_I, p_2, V$ are four parameters of this
model.

We generate 1000 synthetic subjects
(tasks)
following the setup in previous work:
1) the four parameters are sampled from a multivariate
Gaussian with the results from the normal group in Table
1.~of~\citep{vicini2001iterative}, i.e.
\[\begin{split}\boldsymbol\mu &= [2.67, 6.42, 4.82, 1.64]\\ \boldsymbol\Sigma&=\textbf{diag}(1.02, 6.90, 2.34, 0.22);\end{split}\]
2) $I(t)$ is obtained via spline interpolation using the real data
in~\citep{vicini2001iterative}; 3) $G_b$ is fixed to be $84$ and $D$
is set to be 300; 4) $\delta(t)$ is simulated using a Gaussian
profile with its support on the positive axis and the standard
deviation (\emph{SD}) randomly drawn from a uniform distribution on the interval
$[0, 1]$; 5) Noise is added to the observations with $\sigma^2=1$.
Each task has 5 measurements chosen uniformly from the interval $[1,
240]$ and an additional 10 measurements are used for testing. Notice
that the approach in~\citep{pillonetto2010bayesian} cannot deal the
situation efficiently since the inputs do not share samples often.

\begin{figure*}[t]
{
        \begin{minipage}{0.5\textwidth}
        \centering
        \includegraphics[width=\textwidth]{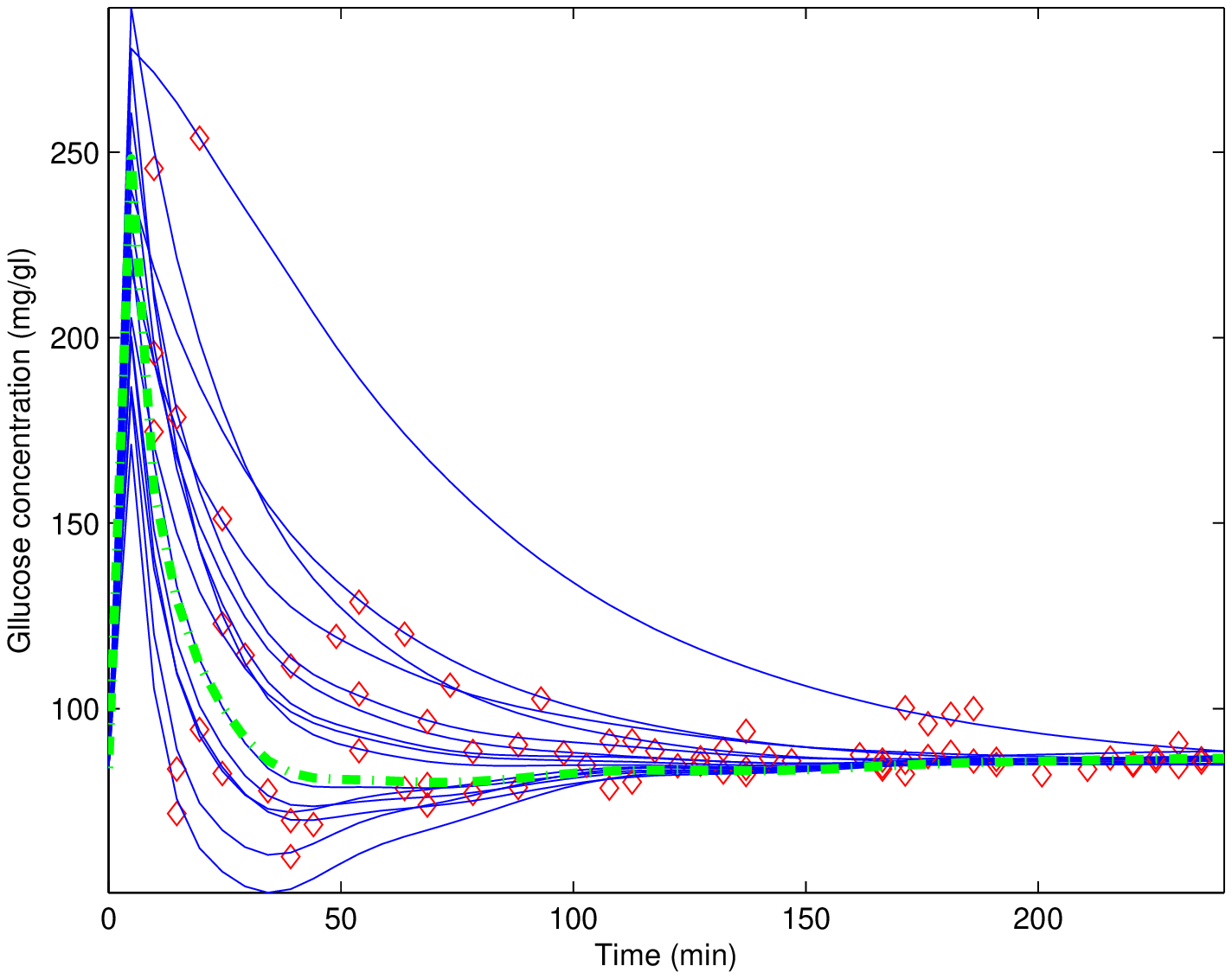}
        \end{minipage}
        \begin{minipage}{0.5\textwidth}
        \centering
        \includegraphics[width=\textwidth]{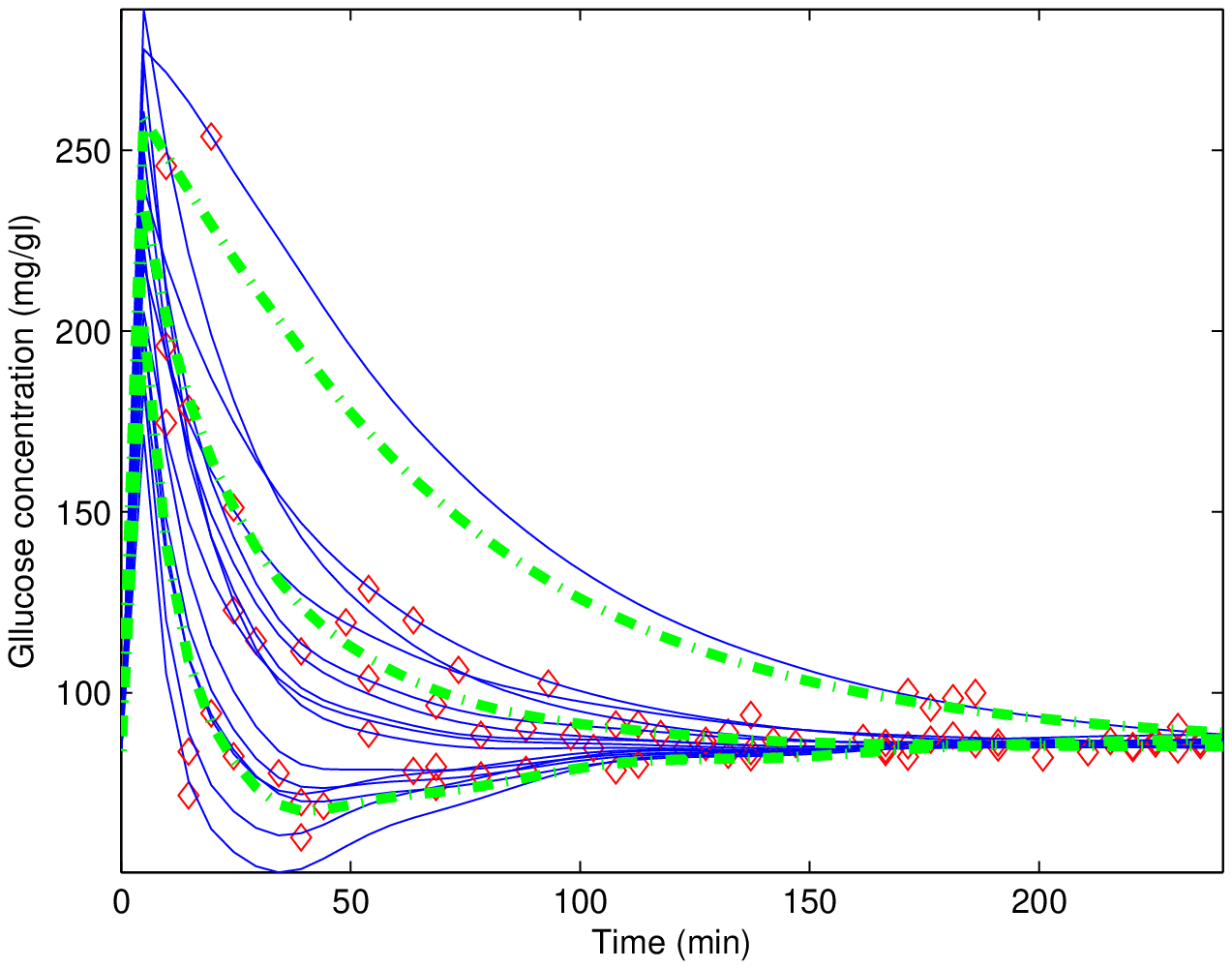}
        \end{minipage}\hfill
        \begin{minipage}{0.5\textwidth}
        \centering
        \includegraphics[width=\textwidth]{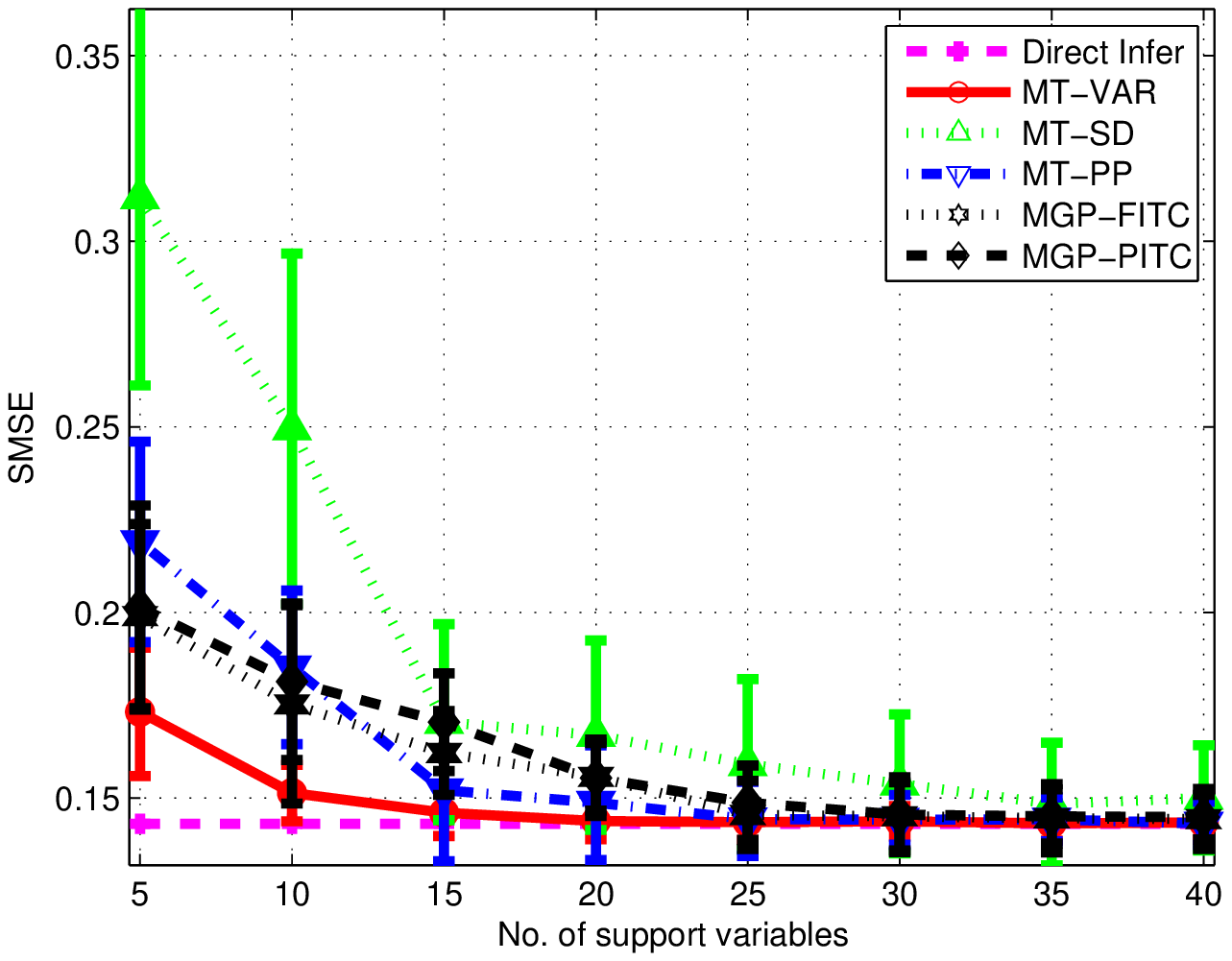}
        \end{minipage}
        \begin{minipage}{0.5\textwidth}
        \centering
        \includegraphics[width=\textwidth]{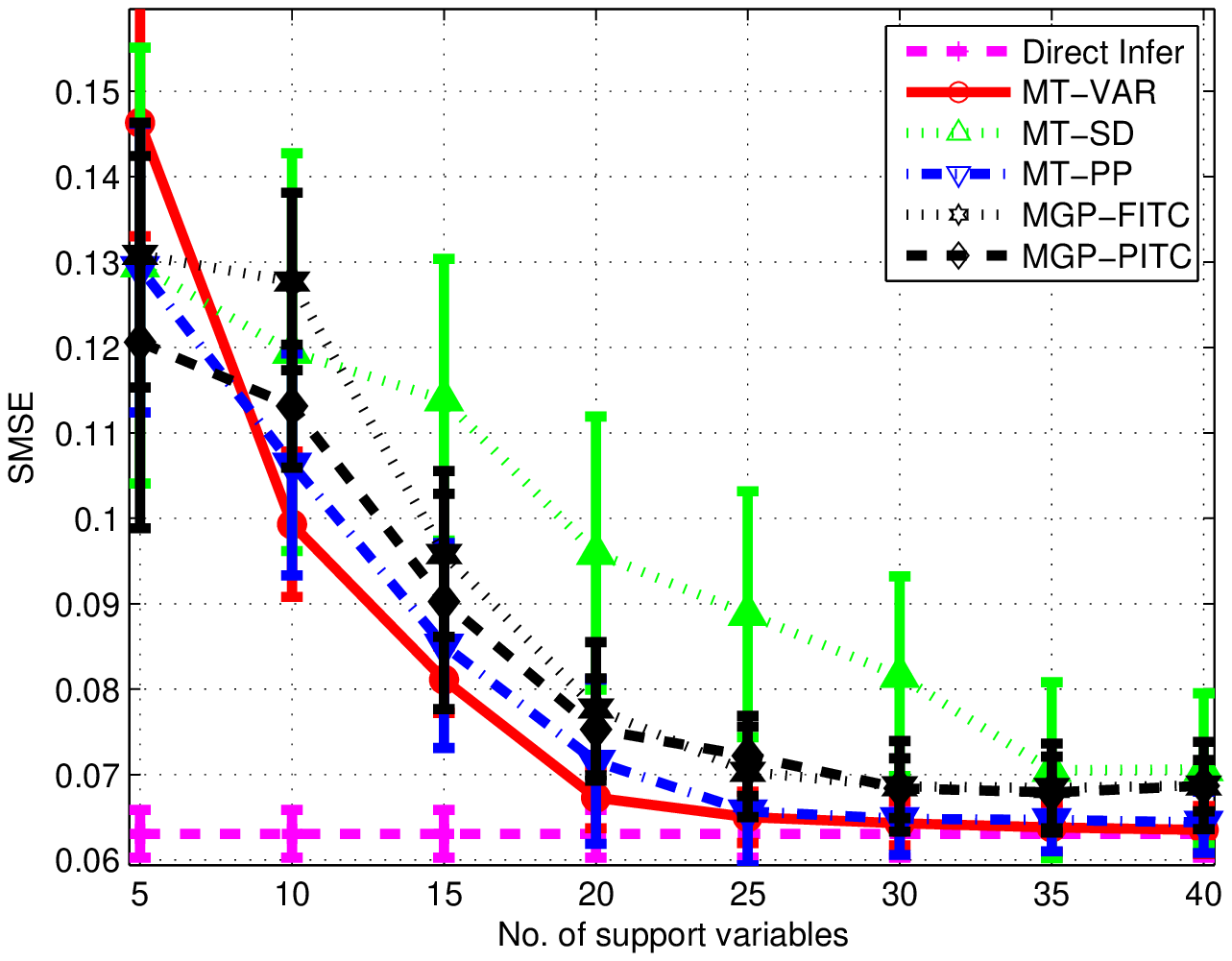}
        \end{minipage}\hfill
        \begin{minipage}{0.5\textwidth}
        \centering
        \includegraphics[width=\textwidth]{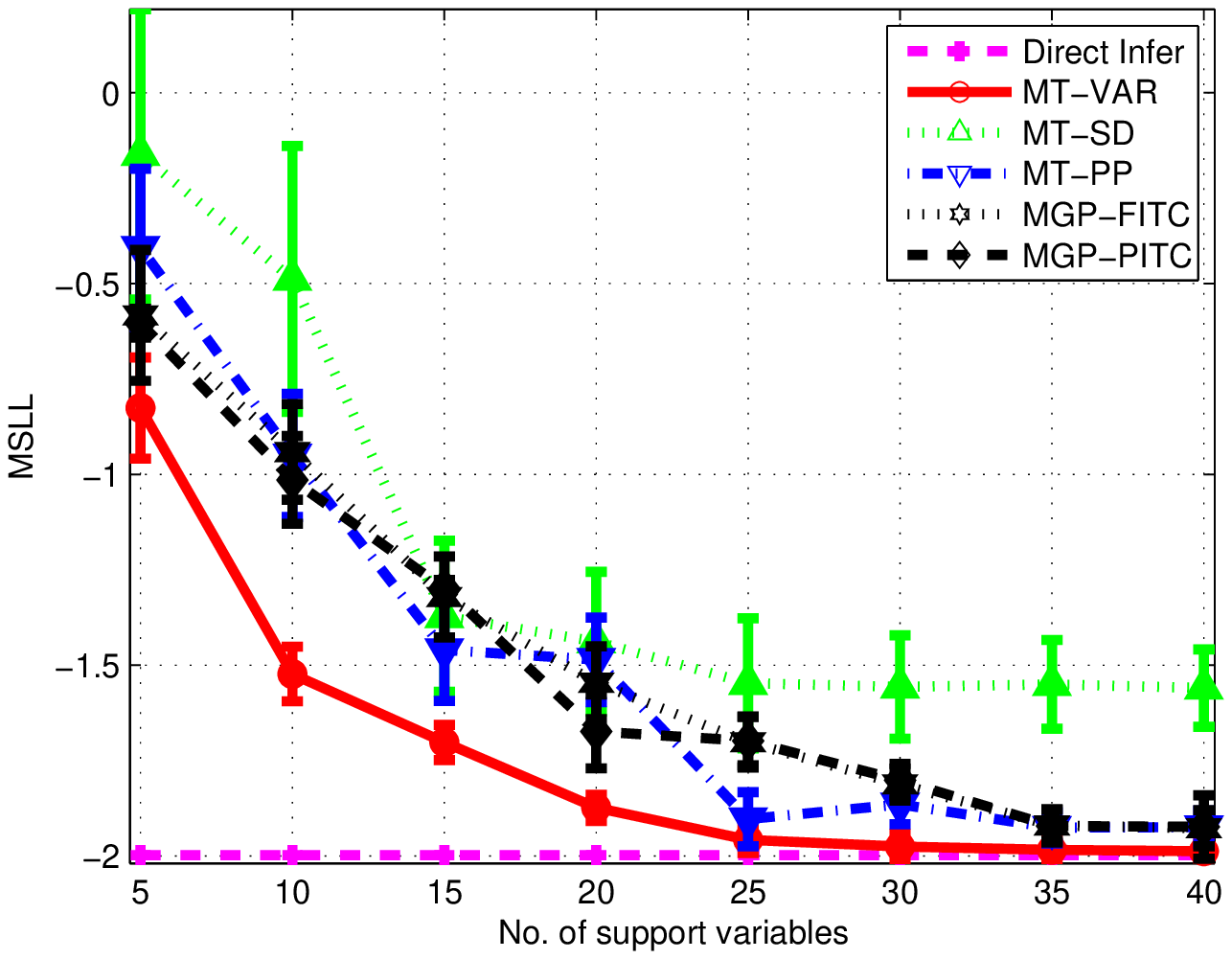}
        \end{minipage}
        \begin{minipage}{0.5\textwidth}
        \centering
        \includegraphics[width=\textwidth]{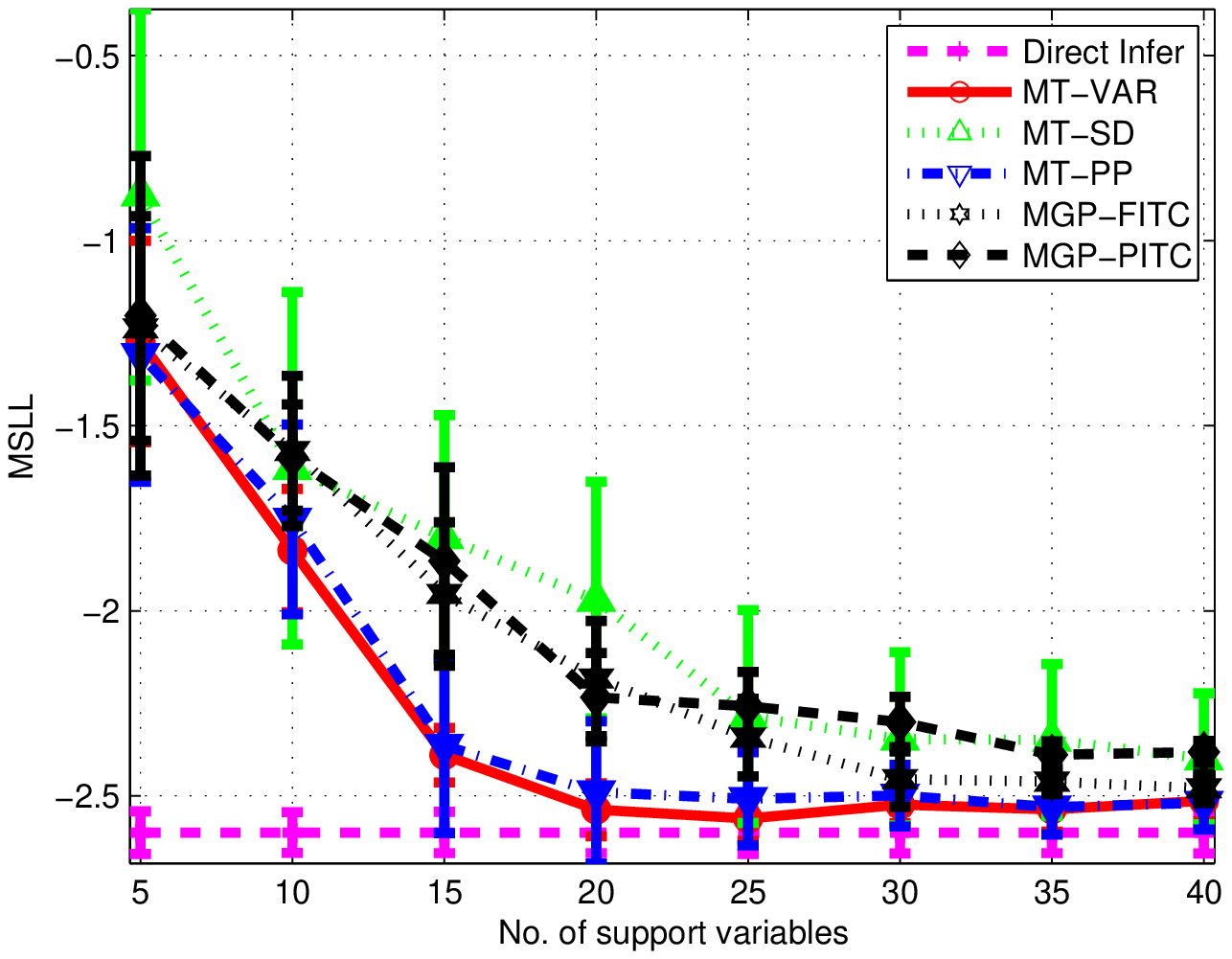}
        \end{minipage}\hfill

\caption{Simulated Glucose Data. Left Column: Single center $K=1$ results; Right Column: Multiple center $K=3$ results;
Top: 15 tasks (Blue) with observations
(Red Diamonds) and estimated fixed-effect curve (Green) obtained
from 1000 IVGTT responses. Although the data is not generated by our
model, it can be seen that different tasks have a common shape and
might be modeled using a fixed effect function plus individual
variations. Middle: The average SMSE for all tasks; Bottom: The
average MSLL for all tasks.} \label{fig:glucose} }
\end{figure*}

The experiments were done under both the single center and the multi center
setting and the results are shown in Fig.~\ref{fig:glucose}. The plots of task distribution on the left
suggest that one can get more accurate estimation by using multiple
centers. For the multiple center case, the number of centers for the proposed
method is arbitrarily set to be 3 ($K=3$) and the number of latent function
of MGP is set to be 2 ($Q=2$) (We were not able of obtain reasonable results
using MGP when $Q=3$). First, we observe that the
multi-center version performs better than the single center one, indicating
that the group-based generalization of the traditional mixed-effect model is
beneficial. Second, we can see that all the methods achieve reasonably good
performance, but that the proposed method significantly outperforms the other
methods.

\subsection{Real Astrophysics Data}
\begin{figure}[H]
{
        \begin{minipage}{0.33\textwidth}
        \centering
        \includegraphics[width=\textwidth]{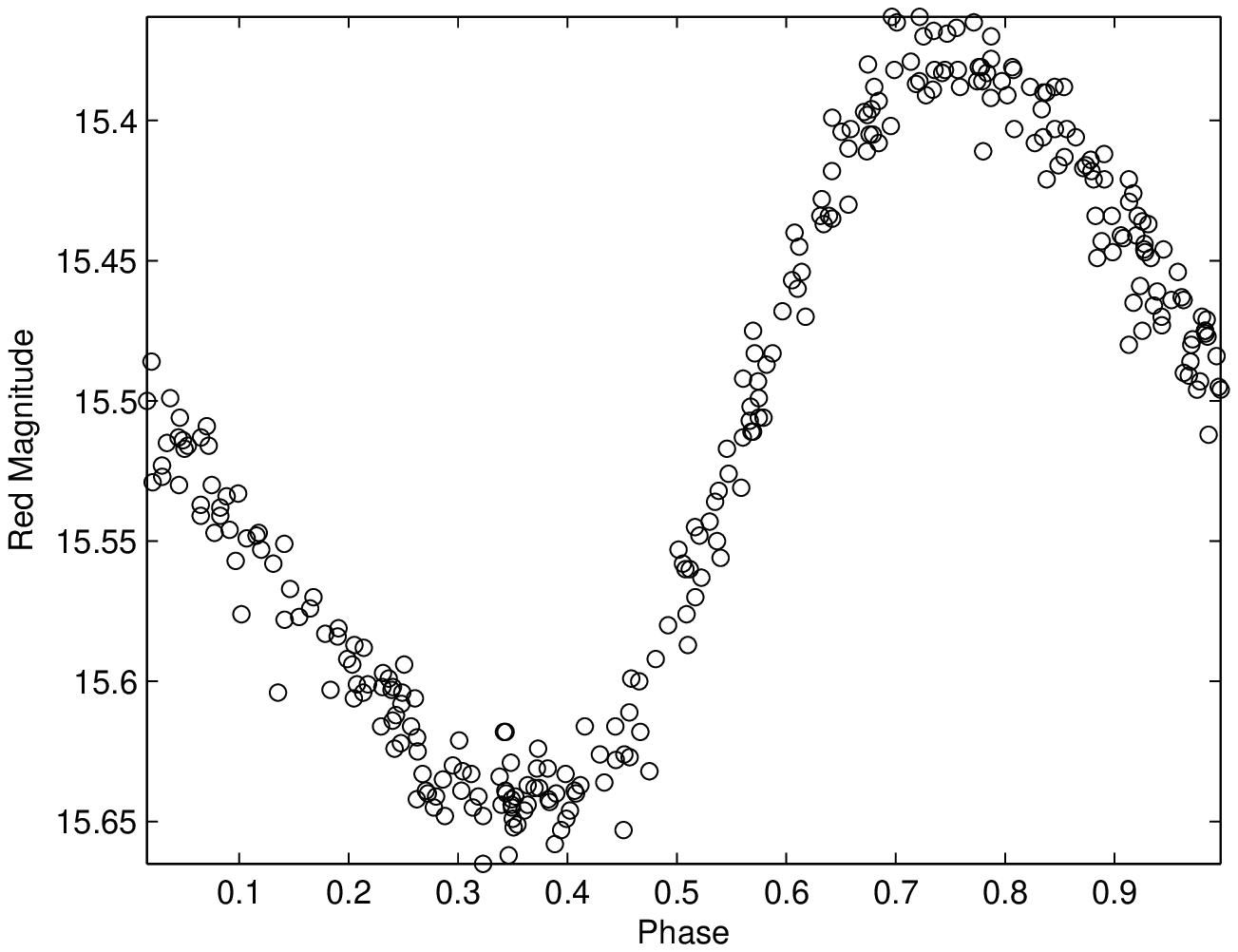}
        \end{minipage}
        \begin{minipage}{0.33\textwidth}
        \centering
        \includegraphics[width=\textwidth]{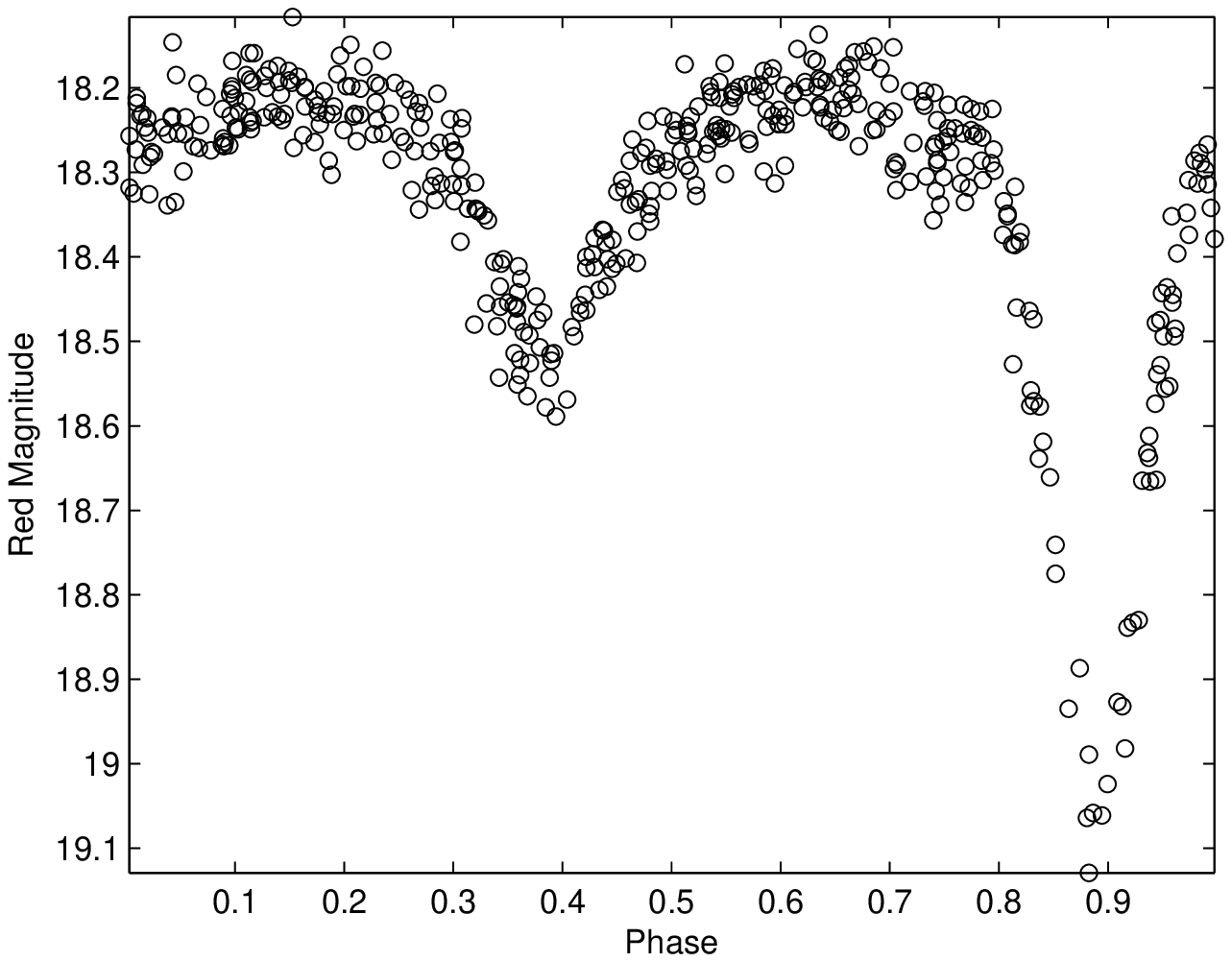}
        \end{minipage}
        \begin{minipage}{0.33\textwidth}
        \centering
        \includegraphics[width=\textwidth]{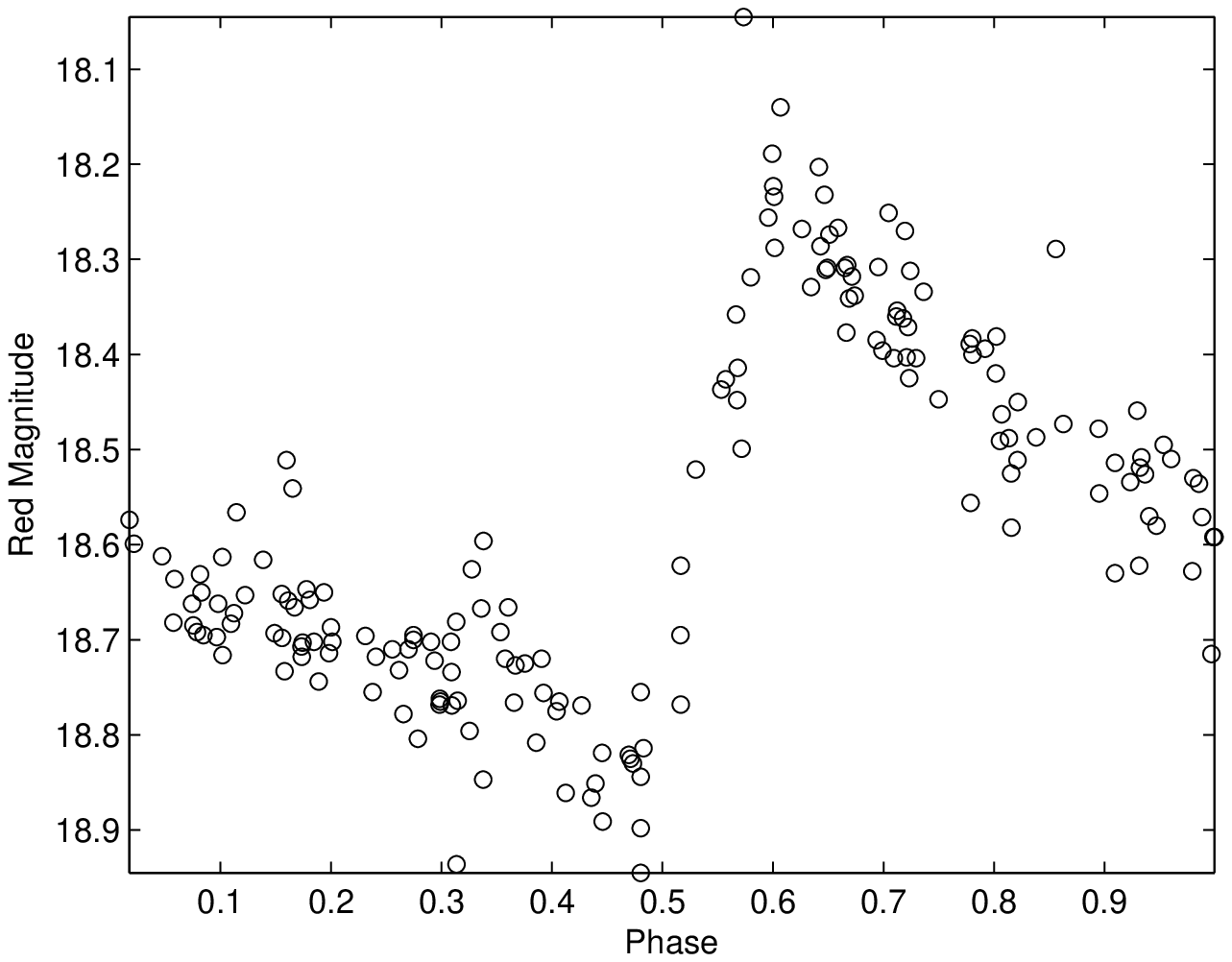}
        \end{minipage}
\caption{OGLEII: time series for one star (one task in our context) of each type.}
\label{fig:oglefold}}
\end{figure}

\begin{figure*}
{
        \begin{minipage}{0.5\textwidth}
        \centering
        \includegraphics[width=\textwidth]{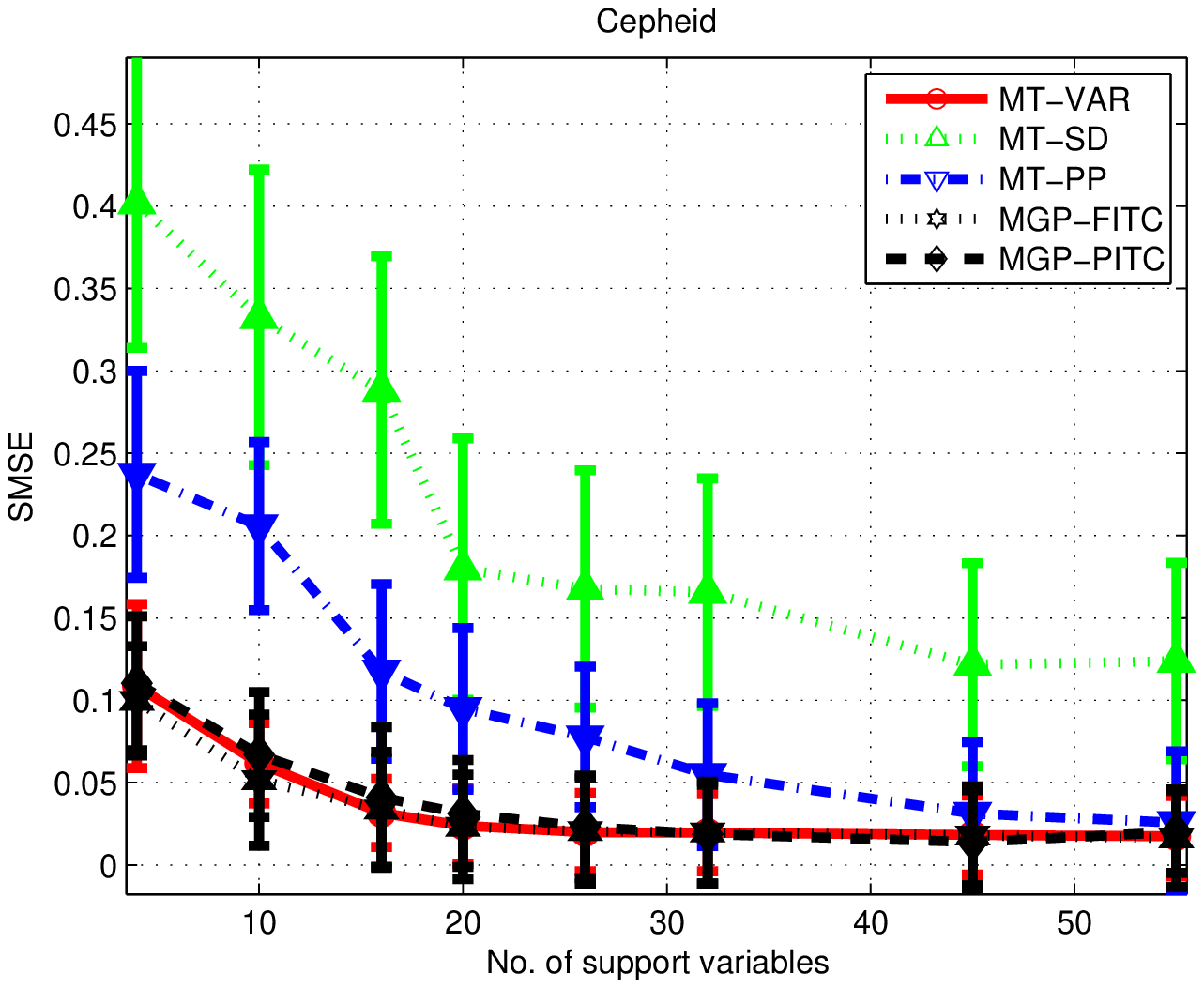}
        \end{minipage}
        \begin{minipage}{0.5\textwidth}
        \centering
        \includegraphics[width=\textwidth]{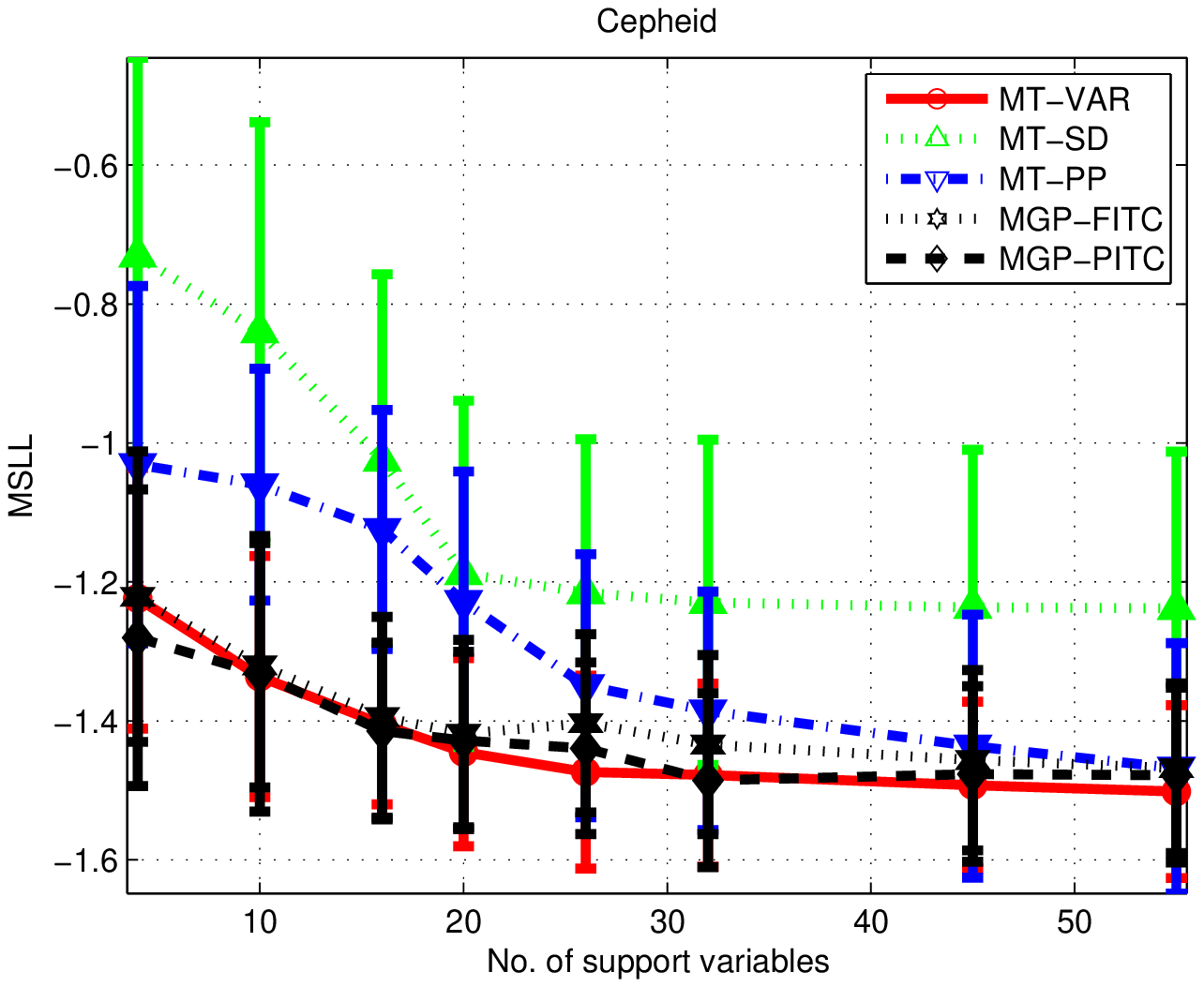}
        \end{minipage}\hfill
        \begin{minipage}{0.5\textwidth}
        \centering
        \includegraphics[width=\textwidth]{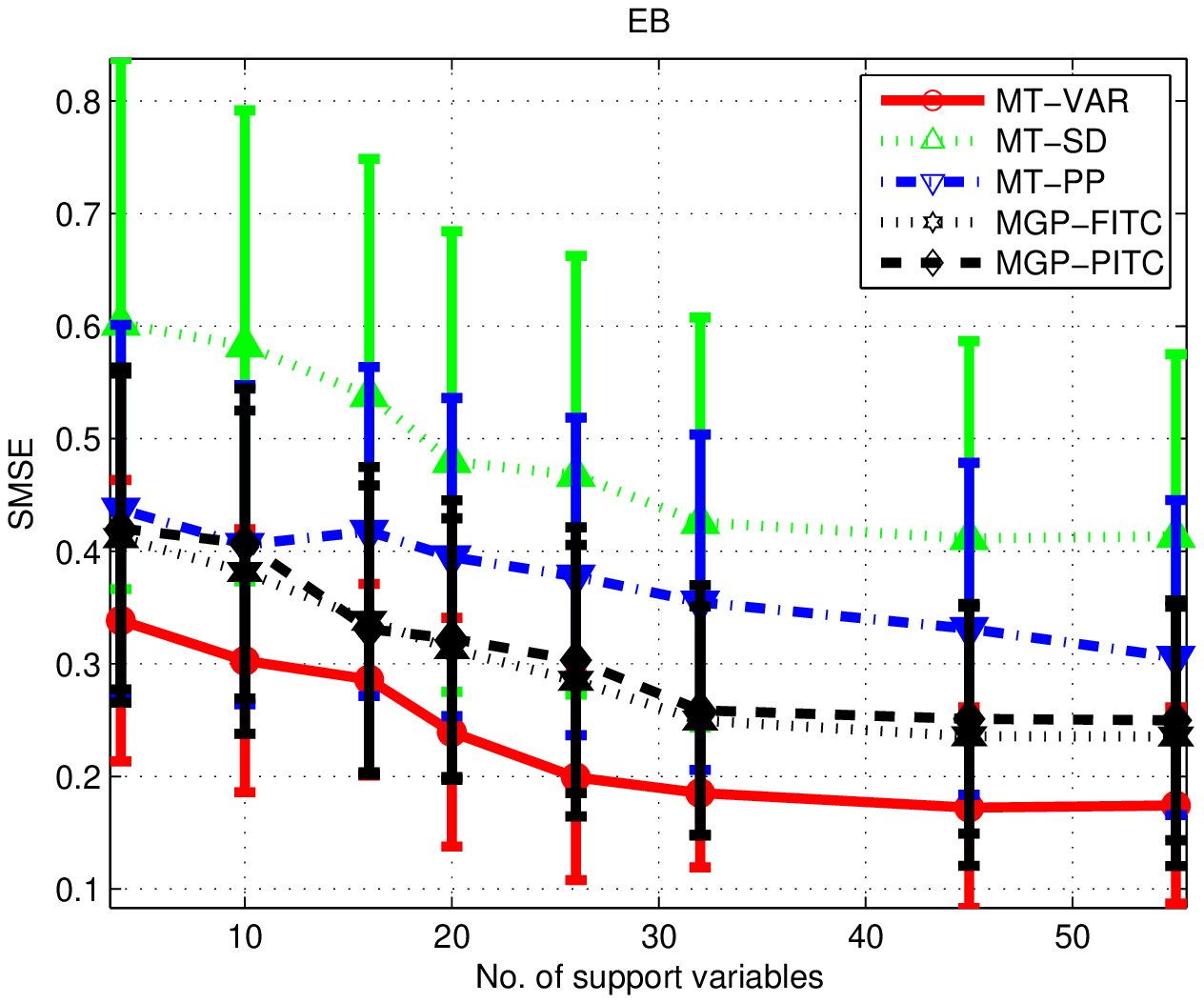}
        \end{minipage}
        \begin{minipage}{0.5\textwidth}
        \centering
        \includegraphics[width=\textwidth]{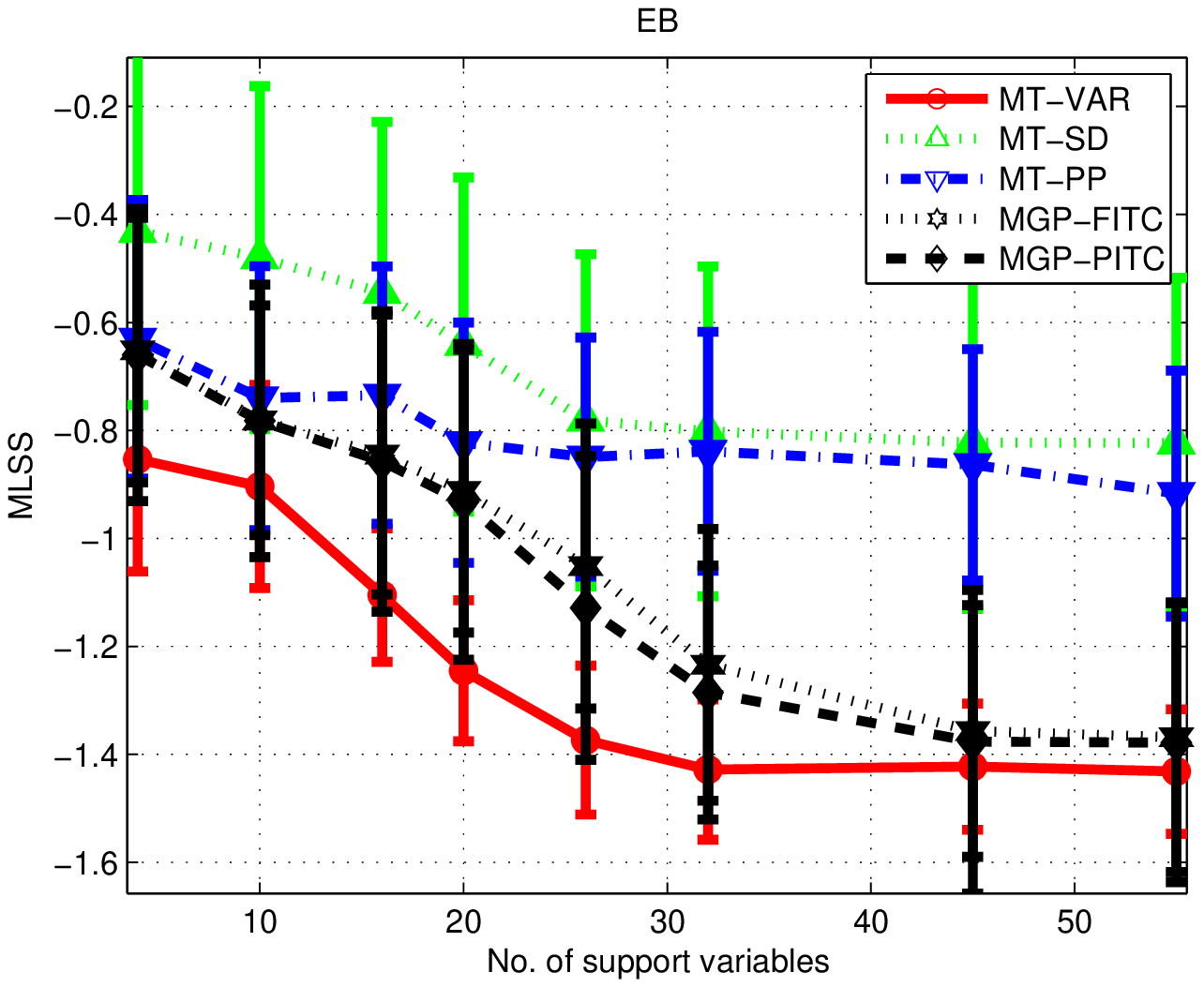}
        \end{minipage}\hfill
        \begin{minipage}{0.5\textwidth}
        \centering
        \includegraphics[width=\textwidth]{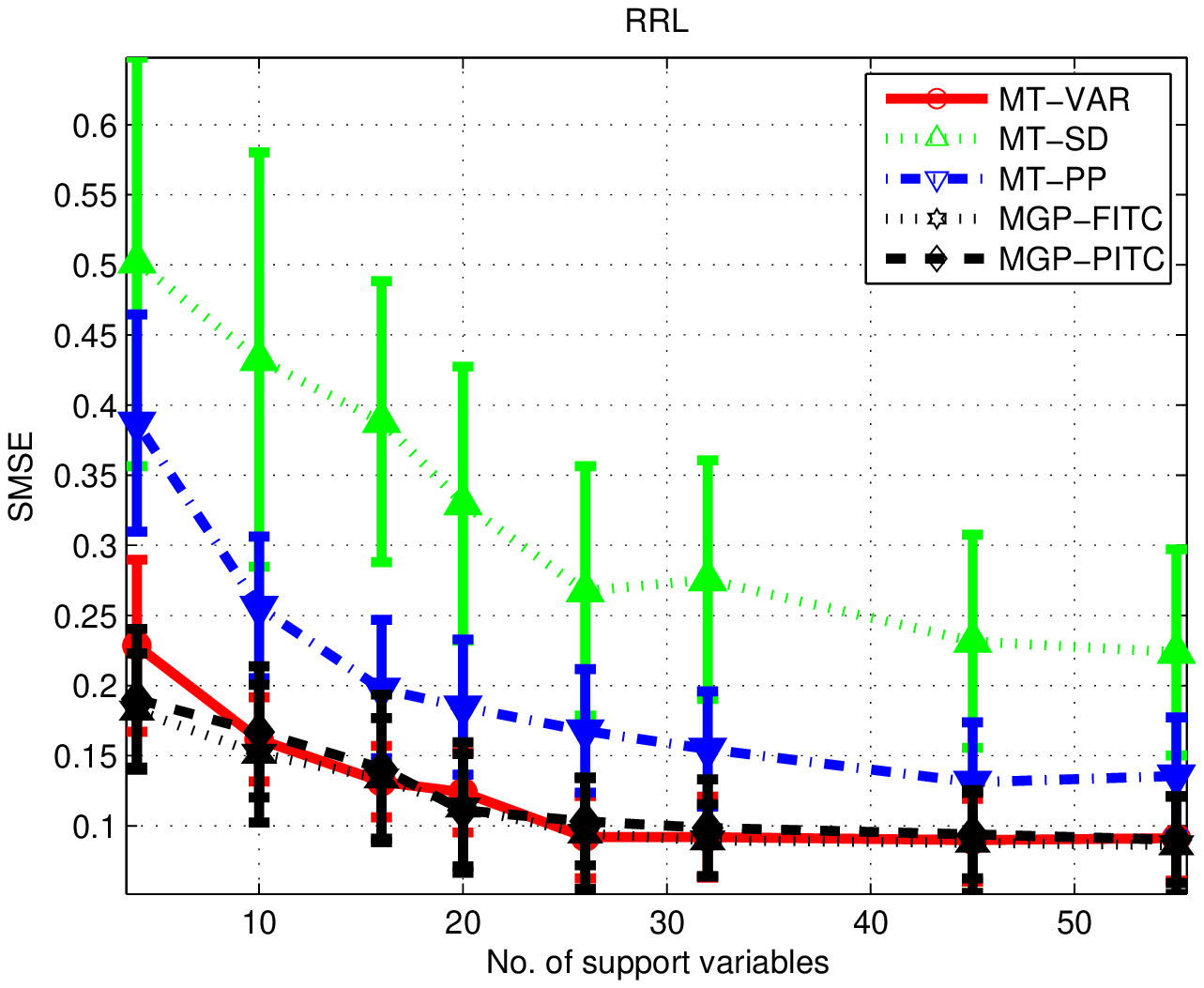}
        \end{minipage}
        \begin{minipage}{0.5\textwidth}
        \centering
        \includegraphics[width=\textwidth]{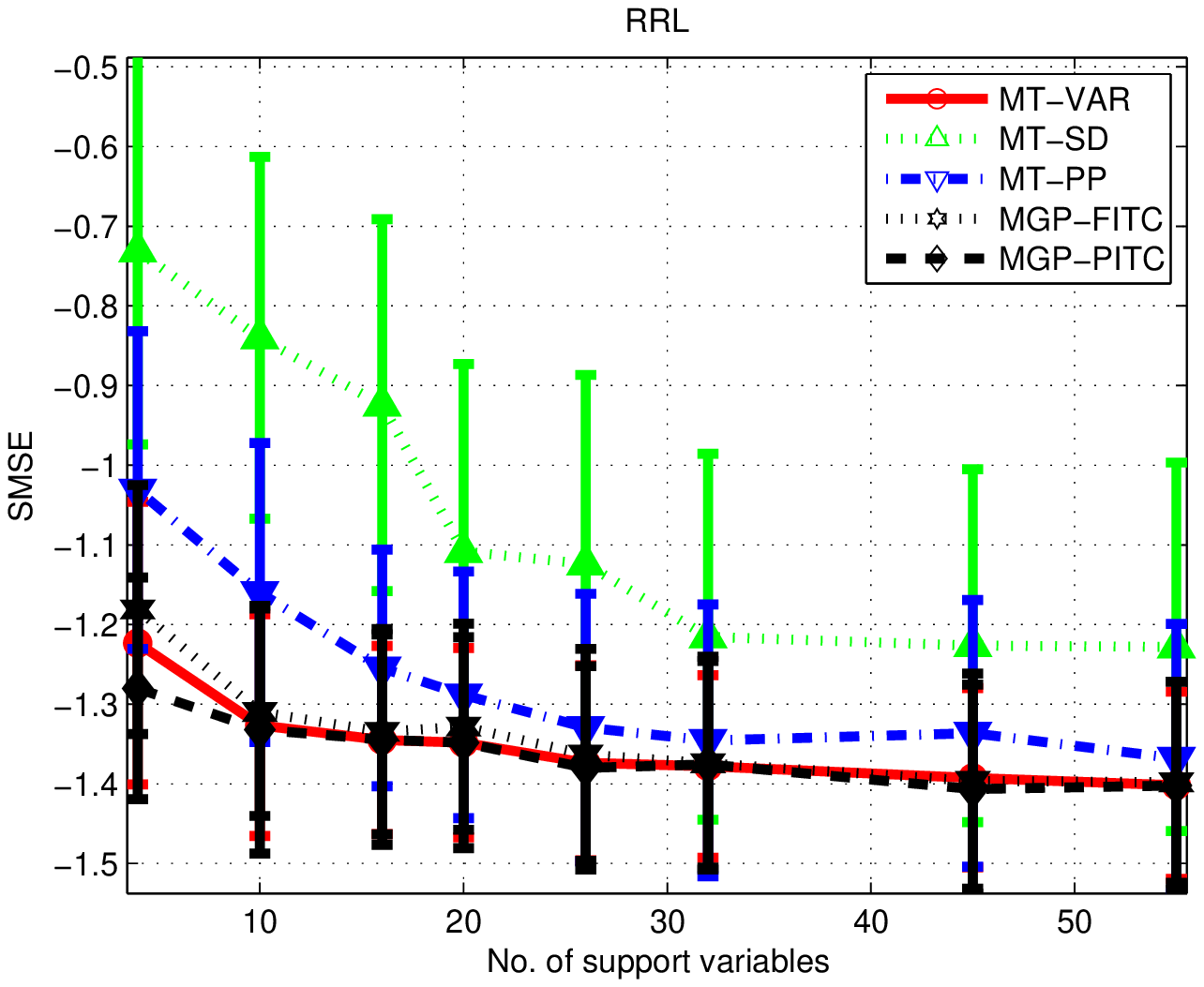}
        \end{minipage}\hfill
        \caption{OGLEII:  The average SMSE and MSLL for all the tasks are shown in the Left and Right Column. Top: Cepheid; Middle: EB; Bottom: RRL.}
\label{fig:ogle}}
\end{figure*}

We evaluate our method using the astronomy dataset of
\citep{wang2010shift}, where a generative model was developed
to capture and classify different types of stars. The dataset,
extracted from the OGLEII survey \citep{soszynski2003optical},
includes stars of 3 types (RRL, CEPH, EB) which constitute 3
datasets in our context.
One example of each class is shown in Fig.~\ref{fig:oglefold}.
These examples are densely sampled but some stars have less
samples and we simulate the sparse case by sub-sampling in our
experiments. In previous work~\citep{wang2010shift}, we developed a
grouped mixed-effect multi-task model that in addition allowed for phase shift of the light
measurements. As shown in~\citep{wang2010shift}, stars of the same type have a range of different shapes and the group structure is useful in modeling this domain. However, for inference,~\cite{wang2010shift} used a simple approach
clipping sample points to a fine grid of 200 equally spaced points,
due to the high dimensionality of the full sample (over 18000
points).
Here we use a random subset of 700 stars (tasks) for each type and
preprocess the data normalizing each star to have mean 0 and standard deviation 1, and using universal phasing~\citep{Protopapas06} to phase
each time series to align the maximum of a sliding window of
$5\%$ of the original points.
For each time series, we randomly sample 10 examples for
training and 10 examples for testing per evaluation of
SMSE and MSLL. The number of centers is set to be 3 for the proposed approach and for MGP we set $Q=1$ (We were not able to use $Q>1$). The results are shown in Fig.~\ref{fig:ogle}. We can see that the proposed model significantly outperforms all other methods on EB. For Cepheid and RRL whose shape is simpler, we see that the error of the proposed model and MGP are very close and both outperform other methods.

\section{Conclusion}
\label{sec:conclu}
The paper develops an efficient variational learning algorithm for the
grouped mixed-effect GP for multi-task learning, which compresses the
information of all tasks into an optimal set of support variables for each
mean effect. Experimental evaluation demonstrates the effectiveness of the
proposed method. In future, it will be interesting to derive an online sparse
learning algorithm for this model. Another important direction is to investigate efficient methods for selection of inducing variables when the input is in high dimensional space. In this case, the clipping method of~\citep{wang2010shift} is clearly not feasible, but the variational procedure can provide appropriate guidance.

\section*{Acknowledgement}
We would like to thank the authors of~\cite{alvarez2011computationally} who kindly made their code available online. This research was partly supported by NSF grant
IIS-0803409. The experiments in this paper were performed on the
the Tufts Linux Research Cluster supported by Tufts UIT
Research Computing.

\bibliography{sparse}
\end{document}